\documentclass{article}

\usepackage{PRIMEarxiv}
\usepackage[utf8]{inputenc} 
\usepackage[T1]{fontenc}    
\usepackage{hyperref}       
\usepackage{url}            
\usepackage{booktabs}       
\usepackage{amsfonts}       
\usepackage{nicefrac}       
\usepackage{microtype}      
\usepackage{lipsum}
\usepackage{fancyhdr}       
\usepackage{graphicx}       
\graphicspath{{media/}}     

\usepackage[utf8]{inputenc} 
\usepackage[T1]{fontenc}    
\usepackage{hyperref}       
\usepackage{url}            
\usepackage{booktabs}       
\usepackage{amsfonts}       
\usepackage{nicefrac}       
\usepackage{microtype}      
\usepackage{lipsum}
\usepackage[numbers]{natbib}
\usepackage{graphicx}
\usepackage{hyperref} 
\usepackage{multirow}
\usepackage{caption}
\usepackage[labelformat=simple]{subcaption}
\usepackage{amsmath,bm}
\usepackage[ruled,vlined]{algorithm2e}
\usepackage{dblfloatfix}
\usepackage{color}

\title{A Comprehensive Survey on Deep Neural Image Deblurring\thanks{This work has been submitted to the Elsevier for possible publication. Copyright may be transferred without notice, after which this version may no longer be accessible.}}

\author{
  Sajjad Amrollahi Biyouki \\
  Department of Industrial and Systems Engineering \\
  The University of Tennessee\\
  Knoxville, TN 37996 \\
  \texttt{samrolla@vols.utk.edu} \\
   \And
 Hoon Hwangbo \\
  Department of Industrial and Systems Engineering \\
  The University of Tennessee\\
  Knoxville, TN 37996 \\
  \texttt{hhwangb1@utk.edu} \\
}
\begin{document}

\maketitle              
\begin{abstract}
Image deblurring tries to eliminate degradation elements of an image causing blurriness and improve the quality of an image for better texture and object visualization.  Traditionally, prior-based optimization approaches predominated in image deblurring, but deep neural networks recently brought a major breakthrough in the field. In this paper, we comprehensively review the recent progress of the deep neural architectures in both blind and non-blind image deblurring. We outline the most popular deep neural network structures used in deblurring applications, describe their strengths and novelties, summarize performance metrics, and introduce broadly used datasets.  In addition, we discuss the current challenges and research gaps in this domain and suggest potential research directions for future works.
\vspace{2mm} 

\noindent{\textbf{Keywords:} blind image deblurring, image restoration, deep neural networks, generative models, convolutional neural networks}

\end{abstract}

\section{Introduction}
\par
Image deblurring, more generally known as an image restoration task, is one of the fundamental techniques in machine learning and image processing that estimates a clearer image from a blurred version \cite{fergus2006removing, krishnan2009fast}. This task improves the texture and quality of images for further usage in machine vision tasks such as object detection and image segmentation. Noisy and atmospheric disturbances, object motion, camera shake, and defocus equipment are common sources creating the degradation of an image. Practically, image deblurring has been applied in a broad range of real-world  applications, including remote sensing \cite{gao2018stacked},\cite{chen2019u}, text documents \cite{hradivs2015convolutional}, face images \cite{shen2018deep}, \cite{lin2020learning}, and generic scenes \cite{nah2017deep}, \cite{tao2018scale}. 
\par
A degraded image ($B$) can be defined as a consequence of the convolution of a clear image ($I$) and a blurriness kernel ($K$) perturbed by additive noise ($n$), which can be formulated as
\begin{equation}
    B = I*K + n,
    \label{blur_eq}
\end{equation}
where $*$ represents the convolution operation. The image deblurring task has two significant taxonomies, namely blind and non-blind deblurring, depending on the information availability of the kernel.  The term ``non-blind'' is used when the kernel is thoroughly or partially known.  On the other hand, when the kernel is totally unknown, both $I$ and $K$ are subject to estimation in Eq.~\eqref{blur_eq}, and such estimation process is called ``blind'' image deblurring.  
In general, blind image deblurring is more challenging, and as a part of the estimation process, it can involve non-blind image deblurring when a kernel estimate becomes available. 
\par
The blind image deblurring model shown in Eq.~\eqref{blur_eq} is essentially ill-posed; that is, numerous combinations of latent images and kernels could be retrieved as a solution.  To address this issue, a prior-based optimization approach, also known as maximum a posteriori (MAP)-based blind image deblurring \cite{li2018learning}, 
tries to solve a regularized problem given as
\begin{equation}
       \min_{I,K} \|I*K -B\|_2^2 + \lambda P(I) + \gamma P(K),
       \label{prior_obj}
\end{equation}
where $\|I*K -B\|$ is a data fidelity term, $P(I)$ and $P(K)$ are priors of a latent image and blur kernel, and $\lambda$ and $\gamma$ are the corresponding regularization parameters, respectively. The type of priors would determine the quality of the retrieved latent image and the way the image's information is extracted.  
Inspired by an early seminal work leveraging a prior for blind image deblurring \cite{fergus2006removing}, various statistical priors have been developed and proposed to form a feasible solution space \cite{cho2009fast, krishnan2011blind, levin2011efficient, sun2013edge, joshi2008psf}. Well-known priors include dark channel prior \cite{pan2016blind}, low-rank prior \cite{ren2016image}, gradient prior \cite{xu2010two}, $l_0$ sparse regularization \cite{xu2013unnatural}, and graph-based prior \cite{bai2018graph}.  While these approaches emphasize the development of more effective priors, some recent studies focus more on shaping the underlying structure of the kernel to regulate the solution space of the kernel directly.  
For instance, \citet{mai2015kernel} fuse the kernel estimates of eight deblurring methods to create a single kernel that can model a complicated structure. In a recent study, \citet{biyouki2021blind} propose a mixture structure of multiple adaptive Gaussian kernels that can theoretically estimate any complex shape of blurriness. Although these prior-based optimization methods led a remarkable advance in image deblurring, the recent advent of learning-based approaches, especially those with deep learning structures, further expedites the advance in image deblurring techniques. 
\par
Neural networks have been applied in various domains, including supply chain management \cite{sharifnia2021robust, bahadoran2022new}, healthcare application \cite{wang2018interactive}, computer vision \cite{minaee2021image}, inverse problems \cite{lucas2018using}, and some image restoration tasks, such as super-resolution \cite{kim2016accurate, dong2014learning, ledig2017photo}, denoising \cite{mao2016image}, and inpainting \cite{yeh2016semantic}.  A very primary work using shallow neural networks to restore a latent image from a blurred image dates back to \citet{steriti1994blind}, but using neural networks for an image deblurring task has not been common until recently.  The recent advance in deep learning has promoted a rapid increase in the usage of various deep neural networks, such as convolution neural network (CNN) \cite{krizhevsky2012imagenet}, recurrent neural network (RNN) \cite{rumelhart1986learning}, and generative network \cite{goodfellow2014generative}, for an image deblurring task.  Such an image deblurring process that involves deep neural networks is often referred to as deep neural image deblurring approaches, which train and learn a mapping function shown as
\begin{equation}
    \hat{I} = F(B, \theta),
\end{equation}
where $\theta$ is a set of network parameters that can be learned from data and $F(\cdot)$ is a restoration function. 
This restoration function is trained to deblur degraded images ($B$) by optimizing the network parameters ($\theta$) based on a selected training loss function. 
\par
The purpose of this paper is to review and highlight the recent development of deep neural networks for image deblurring while focusing on their contributions, deep structure configurations, and popular deblurring mechanisms.  We also discuss future research directions in this field of study. 
Although the deep neural image deblurring approaches gained popularity just recently, there are some other survey papers in the literature.  \citet{su2022survey} outline advances in deep learning structures for general image restoration problems, including image deblurring, denoising, dehazing, and super-resolution. Regarding image deblurring, they briefly describe popular network structures and several well-known deep neural architectures.  Narrowing down to the reviews concerning image deblurring only, \citet{koh2021single} conduct a comparative study of well-known deep neural architectures and categorize the reviewed studies based on their type of deblurring problems, i.e., either blind or non-blind.  From the reviewing perspective, the inclusion of deep neural architectures for non-blind deblurring in addition to those for blind deblurring is one of their major contributions. 
They also present an experiment comparing the performance of the reviewed studies on a new benchmark dataset that has not been used for this purpose. However, only a limited number of works are reviewed in their study.  Likewise, there are a few other survey papers that are not comprehensive. \citet{sahu2019blind} review a few deep neural structures for blind image deblurring and categorize relevant works into two major classes of kernel estimation methods and end-to-end approaches. Another brief survey conducted by \citet{li2022survey} reviews the conventional prior-based optimization methods along with deep neural image deblurring methods.
Meanwhile, as the most recent and significant survey, \citet{zhang2022deep} provide more extensive reviews of deep neural image deblurring approaches. They discuss various blur types, image quality assessment methods, general network architectures with their corresponding loss functions. 
However, since there was a rapid increase in the number of relevant studies conducted after their survey, their review does not involve rich discussion about some recent development, such as deep learning-based image priors and widely used image deblurring mechanisms.  In addition, their categorization and comparisons do not include every paper they reviewed leaving some information missing.  Meanwhile, without describing details of fundamental deep neural architectures, the target audience is limited as most researchers in image deblurring mostly focused on prior-based optimization approaches for a long period of time.

\par
Different from other survey papers, we provide comprehensive reviews of deep neural image deblurring collectively in all aspects stated above.  We first describe the most widely used deep learning architectures and mechanisms in detail to establish knowledge base. Then, we present an extensive survey for both blind and non-blind models while considering unique characteristics of individual studies, their specific deep elements, loss functions, applied datasets, blur types, and usable applications.  
The major contributions of this paper are summarized as follows:
\begin{itemize}
    \item This paper presents details of fundamental deep neural networks broadly used for image deblurring along with their recent developments and advances.  
    \item  This paper provides a comprehensive review of deep neural image deblurring techniques and their deep architectures in both blind and non-blind subcategories and highlights their differences, summarized in tables for clarity. 
    \item This paper surveys deep neural image deblurring models developed for specific applications and deep learning-based image priors that can be adopted in various computer vision tasks, which has not been considered in other survey papers. 
    \item This paper summarizes training loss functions, popular image deblurring datasets, and quantitative performance metrics with their unique specifications and strengths.
    \item This paper collectively presents the performance of all surveyed papers and compares them based on common benchmark datasets and performance metrics to summarize the quality of the proposed approaches.
    \item This paper discusses several challenges and research directions in the deep neural image deblurring field for future works. 
\end{itemize}

\par
The rest of this paper is organized as follows: Sections~\ref{sec:Deep_str} and \ref{sec:Deep_mecha} thoroughly describe popular deep neural network structures and mechanisms used for image deblurring, respectively. In Section~\ref{sec:deep_deblurring}, we extensively review existing studies for generic scene images in terms of their contributions and proposed neural network architectures for both blind and non-blind approaches. In addition, Section~\ref{sec:deep_deblurring} discusses various deep learning-based image priors and image deblurring models developed for specific applications, including face and remote sensing images. Section~\ref{sec:loss_func} describes and compares training loss functions and summarizes their usages based on blur types and applications. Section~\ref{sec:deblur_dataset} outlines the most common and well-known datasets used for image deblurring, and Section~\ref{sec:evaluation} presents popular performance measures and a performance comparison study of the reviewed papers. We discuss some challenges of the current deep neural architectures and provide suggestions for future studies in Section \ref{sec:challenge}, and Section \ref{sec:conclusion} concludes the paper.

\section{Deep Neural Network Structures} \label{sec:Deep_str}
This section presents details of the most common deep network structures used in various computer vision tasks, including image delubrring.  

\subsection{Convolutional Neural Network}
\par
Convolutional neural network (CNN) is one of the structures with significant importance in the domain of deep learning, especially for computer vision tasks \cite{krizhevsky2012imagenet,  kalchbrenner2014convolutional,donahue2014decaf}. \citet{lecun1998gradient} initially proposed the CNN architecture for document recognition to classify two-dimensional data. This architecture adopts two major concepts to improve flexibility in acquiring various shapes with different orientations and distortions in an image: local receptive field and shared weights. A usage of local connections between units in a layer, which was first applied to a visual system \cite{hubel1962receptive}, results in extracting the essential inherent features of the inputs. In the course of multi-layer connections, the most prominent features distinguishing different inputs can be identified. 
In each layer, all such receptive fields share common weights, reducing the computational cost dramatically by estimating far fewer weights than a fully-connected network requires \cite{minaee2021image}.
\begin{figure}[!h]
    \centering
    \includegraphics[width=0.9\textwidth]{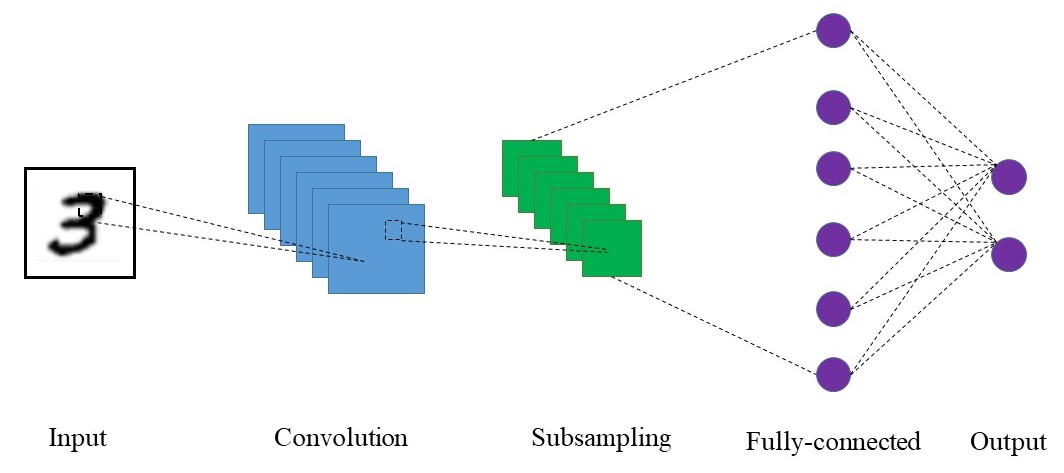}
    \caption{A common CNN architecture}
    \label{fig:CNN}
\end{figure}
\par
Some specific types of layers have been widely used to construct a CNN.  Convolutional layers extract the major structure of the previous layer by convolving a filter, also known as a kernel, with neighboring pixel values.  Subsampling (pooling) layers diminish the dimension of feature maps by applying statistical operators, such as average or max, to small blocks of neighboring pixels. Then, fully-connected layers connect all input neurons of one layer with each neuron in the next layer. 
These fully-connected layers have been extensively used in conventional feed-forwarding neural networks for supervised learning tasks. 
Fig.~\ref{fig:CNN} illustrates a common structure of CNN with the aforementioned layers.
\par
Various CNN architectures with deeper structures and larger receptive fields have been developed to improve the network's performance while managing the computational cost. Well-known CNN structures include AlexNet \cite{krizhevsky2012imagenet}, VGGNet \cite{simonyan2014very}, GoogleNet \cite{szegedy2015going}, and Residual Network (ResNet) \cite{he2016deep}. ResNet\cite{he2016deep} is one of the most applied and well-known architectures in the deep neural image deblurring field, which is explained in detail separately in the following section. 

\subsection{Residual Network}
\par
\begin{figure}[!h]
    \centering
    \includegraphics[width=0.7\textwidth]{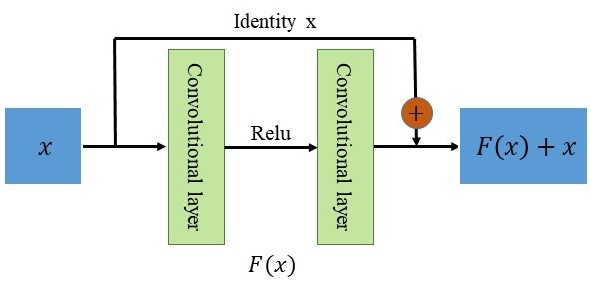}
    \caption{Residual block structure}
    \label{fig:Resnet}
\end{figure}

As a network becomes deeper with a large number of layers, a degradation problem, also known as the gradient vanishing problem \cite{nielsen2015neural}, occurs. 
ResNet was introduced to address this degradation problem for more effective construction of deeper networks  \cite{he2016deep}.  This problem specifically refers to a situation where network performance does not improve anymore but starts degrading as a network gets deeper \cite{he2016deep}. This is because gradient information for updating weights becomes trivial in deeper layers, and the corresponding weights are hardly updated in a back-propagation process.  To tackle this problem, \citet{he2016deep} proposed to propagate the information of an earlier layer directly into deeper ones while skipping some intermediate layers by applying a residual network architecture.  As shown in Fig.~\ref{fig:Resnet}, this architecture consists of several residual blocks that are designed to extract features present in the residuals of the original information.  In this way, high-level features can readily pass through the network without experiencing the gradient vanishing problem.   

\subsection{Encoder-Decoder Network}
\par
\begin{figure}[!h]
    \centering
    \includegraphics[width=0.95\textwidth]{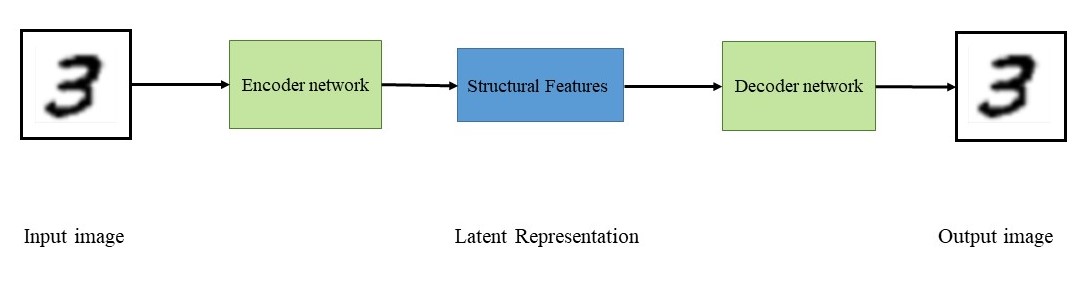}
    \caption{Auto-encoder network}
    \label{fig:autoencoder}
\end{figure}

Encoder-decoder networks are a family of symmetric CNN structures that seek to learn a latent space representing the most prominent features of the input \cite{mao2016image}. The encoder's task is to map data into latent spaces with lower dimensions, and the decoder learns to estimate the output based on the features defined in the latent space. Autoencoders \cite{goodfellow2016deep, hinton2006reducing} are one of the most widely used structures, of which inputs and outputs are set to be the same. In the training procedure, their loss function is typically defined based on the difference between the original output and reconstructed output which needs to be minimized. 
Fig.~\ref{fig:autoencoder} illustrates the general structure of an autoencoder model. 
\par
Inspired by the success of autoencoders, several models enhancing the vanilla autoencoder have been proposed.  A denoising autoencoder uses a corrupted noisy input instead of a clear one to enforce a model to extract the original structure of the input more effectively \cite{vincent2008extracting}. This type of autoencoder tries to learn from a degraded input and reconstruct a undistorted output in order to undo the noisy effects, which is more complicated than the vanilla autoencoder that keeps the inputs and outputs the same. Variational autoencoders (VAE) \cite{kingma2013auto} are another promising structure in the family of generative encoder-decoder models, and they represent the latent space through a distribution of the input's substantial features. This technique trains hyperparameters of the distribution and samples data from the distribution to generate new data as an output. 

\subsection{Sequential-based Networks}
\par
Sequential networks are commonly used for tasks that involve sequenced data, such as speech recognition, natural language processing, and time-series prediction \cite{liu2020dstp, salehinejad2017recent}.  For this network structure, either input or output, or both, could be sequential. Some major sequential networks are described in the following sections.      
 
\subsubsection{Recurrent Neural Network}
\par
\begin{figure}[!h]
    \centering
    \includegraphics[width=0.9\textwidth]{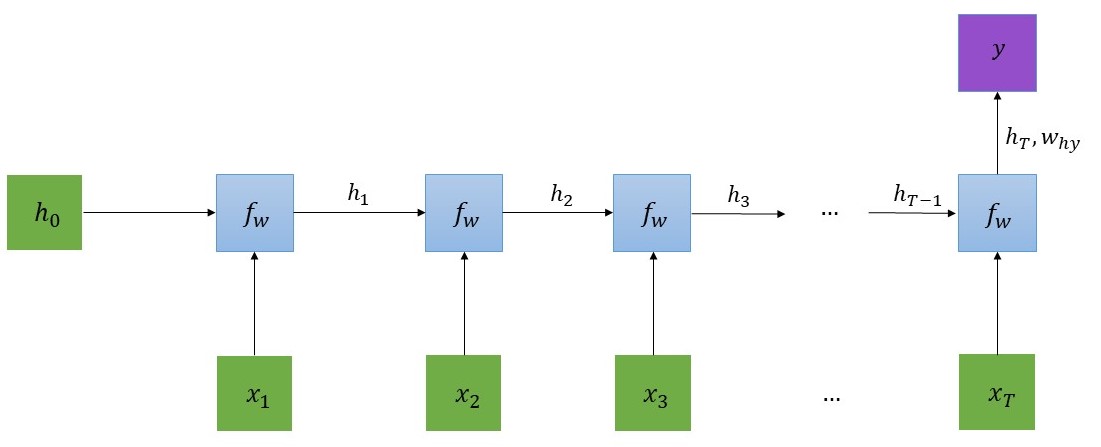}
    \caption{Unroll recurrent neural network}
    \label{fig:RNN}
\end{figure}

Recurrent neural networks (RNNs) \cite{rumelhart1986learning, hopfield1982neural} have an internal loop state to maintain information while processing sequential data \cite{lipton2015critical}. The input of a network at time $t$ ($x_t$) and previous hidden state at time $t-1$ ($h_{t-1}$) are fed into a recurrent neuron defined for the current timestamp ($t$) which returns an output ($y_t$) as well as a hidden state of the network ($h_t$) at time $t$. Figure \ref{fig:RNN} illustrates the general structure of RNN, which can be modeled by
\begin{equation}
    h_t = f_w(h_{t-1}, x_t)
\end{equation}
\begin{equation}
    y_t = w_{hy} \cdot h_t
\end{equation}
where $f_w(\cdot)$ and $w_{hy}$ are an adopted activation function ($tanh$ function in most cases) and the output weights, respectively. It is worth noting that the weights are the same for all time steps, so they are independent of the time sequence. 
\par
A major shortcoming of RNNs is that they cannot accurately infer long-term dependencies. In other words, if the output at the current timestamp depends on information that was available a long time ago, the network cannot make a connection between the required information and the output. In addition, in the RNN structure, the gradient vanishing problem occurs quite often. These challenges can be addressed by defining some interacting layers as in long short-term memory (LSTM) networks \cite{hochreiter1997long} shown in detail in the next section. 

\subsubsection{Long Short-Term Memory Network}
\par
\begin{figure}[!h]
    \centering
    \includegraphics[width=0.9\textwidth]{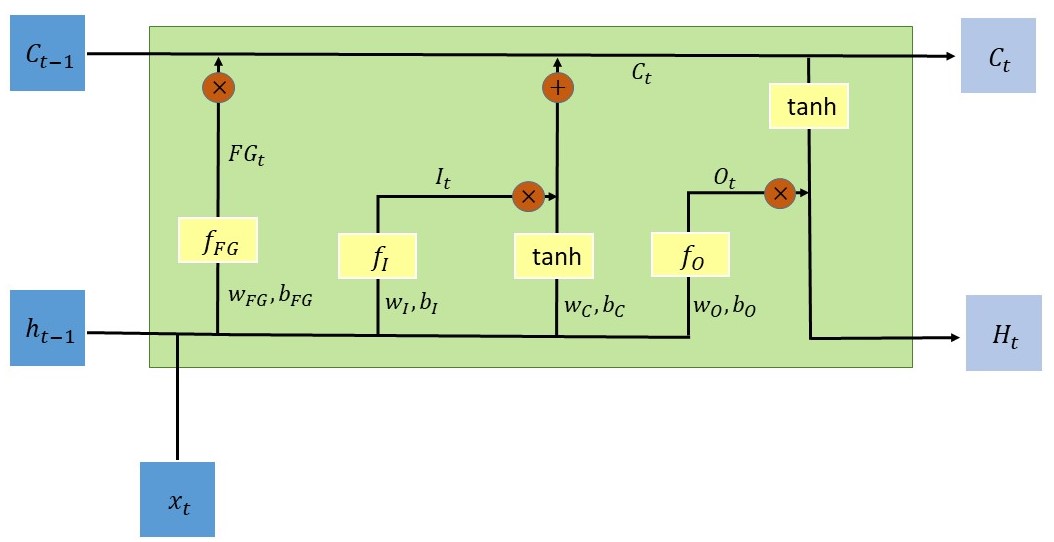}
    \caption{Long short-term memory architecture}
    \label{fig:LSTM}
\end{figure}

\citet{hochreiter1997long} proposed an LSTM network to improve the RNN structure by retrieving long-term information and diminishing the effects of the gradient vanishing problem. To achieve this, an LSTM network consists of a forget gate layer (Eq.~\eqref{f_input}), an input gate layer (Eqs.~\eqref{I_input}-\eqref{I_input2}), and an output gate layer (Eqs.~\eqref{output}-\eqref{output2}), all of which remember some long-period information and regulate the input-output information in its cell state \cite{minaee2021image}. Concerning the formulas, $FG_{\hspace{1mm} t}$ is the forget gate which stores part of the current input data ($x_t$) along with the previous hidden state ($h_{t-1}$). $I_t$ is the input gate that decides how much information gets through, and the output gate, $O_t$, extracts a part of the input information ($x_t$ and $h_{t-1}$) to be transferred to the next LSTM unit. $f_i(\cdot)$ for $i \in \{FG, I, O\}$ are activation functions for the three gates (mostly sigmoid function) and $C_t$ and $H_t$, respectively, are the ultimate output and current hidden state that will be transferred to the next time step. Figure \ref{fig:LSTM} illustrates the structure of the LSTM layers. 
\begin{equation}
    FG_{\hspace{1mm} t} = f_{FG}(w_{FG}.[h_{t-1},x_t]+b_{FG})
    \label{f_input}
\end{equation}

\begin{equation}
    I_t = f_I(w_I.[h_{t-1},x_t] + b_I)
    \label{I_input}
\end{equation}

\begin{equation}
    C_t = FG_{\hspace{1mm} t}.C_{t-1} + I_t.\tanh{(w_C. [h_{t-1},x_t] + b_C)}
    \label{I_input2}
\end{equation}

\begin{equation}
    O_t = f_O(w_O[h_{t-1}, x_t] + b_O)
    \label{output}
\end{equation}

\begin{equation}
    H_t = O_t.\tanh{(C_t)}
    \label{output2}
\end{equation}

\subsection{Generative Adversarial Networks}
\par
\begin{figure}[!h]
    \centering
    \includegraphics[width=0.8\textwidth]{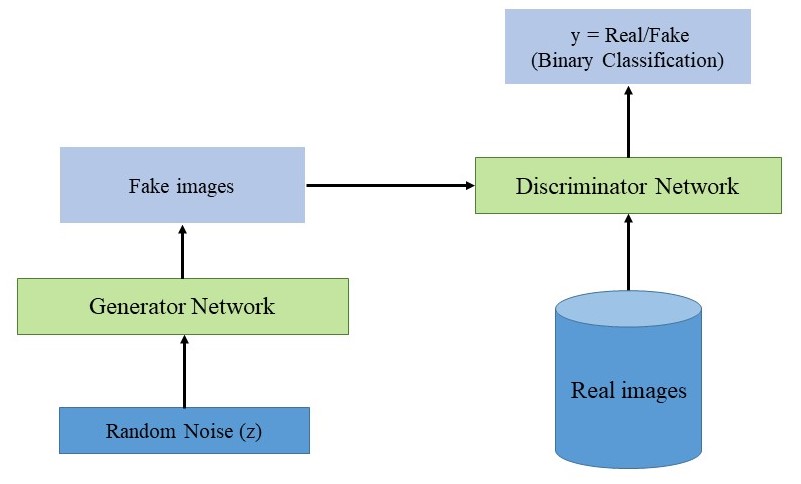}
    \caption{GAN architecture}
    \label{GAN Architecture}
\end{figure}

In general, generative models generate applicable samples by learning a data generation process and its corresponding distribution, especially when data reside in a high-dimensional space \cite{tu2007learning}. In the deep learning context, generative adversarial networks (GANs) \cite{goodfellow2014generative} are one of the most well-known generative models. GANs learn a mapping function that transforms a simple random distribution to the data distribution, allowing it to be used to generate samples. As shown in Fig.~\ref{GAN Architecture}, their architecture is built upon two fundamental networks: generator network and discriminator network. A generator network tries to generate fake but realistic images for distraction, and a discriminator network aims to discern real images from the artificial fake images. 
\par
An optimization process is used to train these networks simultaneously through a minmax loss function defined as
\begin{equation}
    \min_{G}\max_{D} [E_{I\sim p_{r}(I)} \text{log}(D(I)) + E_{z\sim p_{z}(z)} log(1-D(G(z)))] 
    \label{GAN}
\end{equation}
where $p_{r}$ and $p_z$ are the real data distribution and a noise distribution, respectively. In Eq.~\eqref{GAN}, the discriminator aims to maximize the objective function in a way that $D(I)$ and $D(G(z))$ get close to one for a real image and zero for a fake generated image, respectively. On the other hand, the generator tries to minimize the entire objective function by having the $D(G(z))$ value close to one. A major goal of employing the discriminator is to enforce the generator distinguish the characteristics of real images against similar-looking fake images. With this unique benefit of a GAN structure, various adjustments to the GAN structure have been made to enhance its performance, overcome training difficulties, and alleviate its computational cost.  This includes Wesserstein-GANs   \cite{arjovsky2017wasserstein, gulrajani2017improved}, conditional-GANs \cite{mirza2014conditional}, least squares-GANs \cite{mao2017least} and Markovian-GANs \cite{li2016precomputed}.  
\par
The conditional-GAN (CGAN) is commonly used when both generator and discriminator networks can be conditioned on additional information ($\psi$) such as class labels and data from other modalities \cite{mirza2014conditional}. The loss function of CGAN is formulated as
\begin{equation}
    \min_{G}\max_{D} [E_{I\sim p_{r}(I)} \text{log}(D(I|\psi)) + E_{z\sim p_{z}(z)} log(1-D(G(z|\psi)))].
    \label{CGAN}
\end{equation}
where both generator and discriminator networks are conditioned by additional input layer $\psi$.
\par
Often, the original GAN structure cannot be learned effectively due to mode collapse and gradient vanishing problems \cite{kupyn2018deblurgan, salimans2016improved}. To overcome these problems and improve the training process, Wassertein GAN (WGAN) \cite{arjovsky2017wasserstein} was proposed. WGAN adopts Wasserstein-1 distance to minimize the divergence of the distributions instead of Jensen-Shannon approximation used in the vanilla GAN \cite{kupyn2018deblurgan}. In WGAN, a loss function is modeled as
\begin{equation}
    \min_{G}\max_{D \in \mathcal{D}} E_{I\sim p_r(I)}[D(I)] - E_{I'\sim p_g(I')} [D(I')] 
    \label{WGAN}
\end{equation}
where $\mathcal{D}$ is the set of 1-Lipschitz functions, $I'$ is the generated image from a noise distribution ($p_z$), $p_g$ and $p_r$ are the generated and real data distributions, respectively. The discriminator network $D$, also called critic, approximates the Wassertein distance between $p_g$ and $p_r$ as $K \cdot W(p_g,p_r)$, where $K$ is a Lipschitz constant. Therefore, the discriminator network's weights are truncated to the range of $[-c,c]$ where $c$ is a positive constant, to enforce the Lipschitz constraint on the discriminator of WGAN  \cite{arjovsky2017wasserstein}. As an alternative technique enforcing the Lipschitz constraint,  a gradient penalty term can be added to the WGAN loss function in Eq.~\eqref{WGAN} \cite{gulrajani2017improved}, which is formulated as
\begin{equation}
    \min_{G}\max_{D \in \mathcal{D}} E_{I\sim p_r(I)}[D(I)] - E_{I'\sim p_g(I')} [D(I')] + \lambda  E_{\tilde{I} \sim p_{\tilde{I}}(\tilde{I})} [(\| \nabla_{\tilde{I}} D(\tilde{I})\|_2 - 1)^2]
    \label{im_WGAN}
\end{equation}
where $p_{\tilde{I}}$ is the penalty distribution computed between $p_{g}$ and $p_{r}$, and $\tilde{I}$ are random samples generated from $p_{\tilde{I}}$. It has been shown that this additional regularization term can improve the stability of training as well as the performance of a GAN model \cite{zhao2021gradient}.

\section{Prominent Mechanism in Neural Image Deblurring}
\label{sec:Deep_mecha}
This section describes some mechanisms widely used in deep neural image deblurring specifically, such as skip connection, pyramid scheme, and attention.  
\subsection{Skip Connection}
\par
\begin{figure}[!h]
    \centering
    \includegraphics[width=0.9\textwidth]{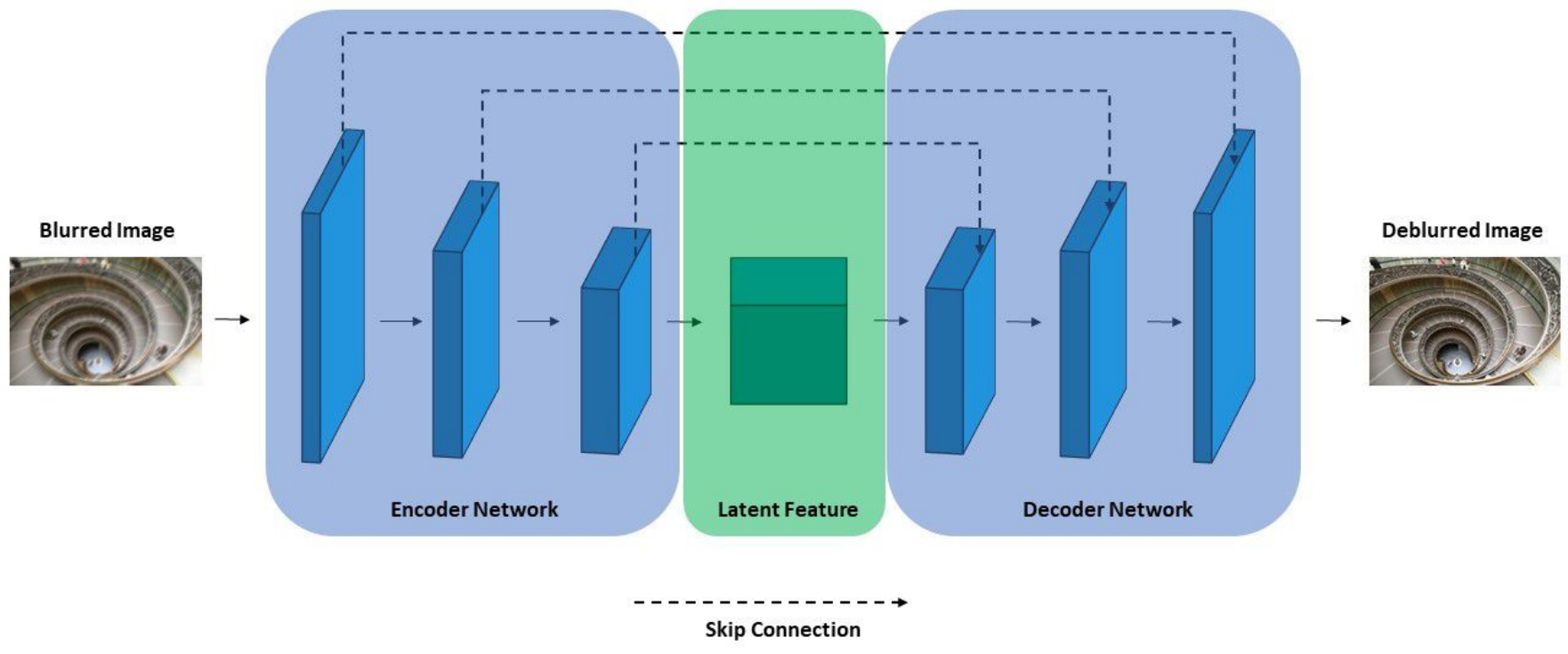}
    \caption{Skip connection in an encoder-decoder diagram}
    \label{fig:SkipConn}
\end{figure}

As shown in Fig.~\ref{fig:SkipConn}, skip connection \cite{long2015fully} uses hierarchical features through all the convolutional layers, which was proposed in the U-Net architecture \cite{ronneberger2015u} that has a symmetric encoder-decoder framework. In the field of image deblurring, skip connection effectively captures different levels of blurry features in the layers \cite{zhao2021gradient}, and it can boost convergence and gradient propagation \cite{tao2018scale}. In this survey paper, the skip connection mechanism is regarded as multiple connections in the encoder-decoder structure (same as the U-Net structure) rather than a global connection as in ResNet blocks. 

\subsection{Multi-scale (Pyramid) Scheme}
\par
\begin{figure}[!h]
    \centering
    \includegraphics[width=0.9\textwidth]{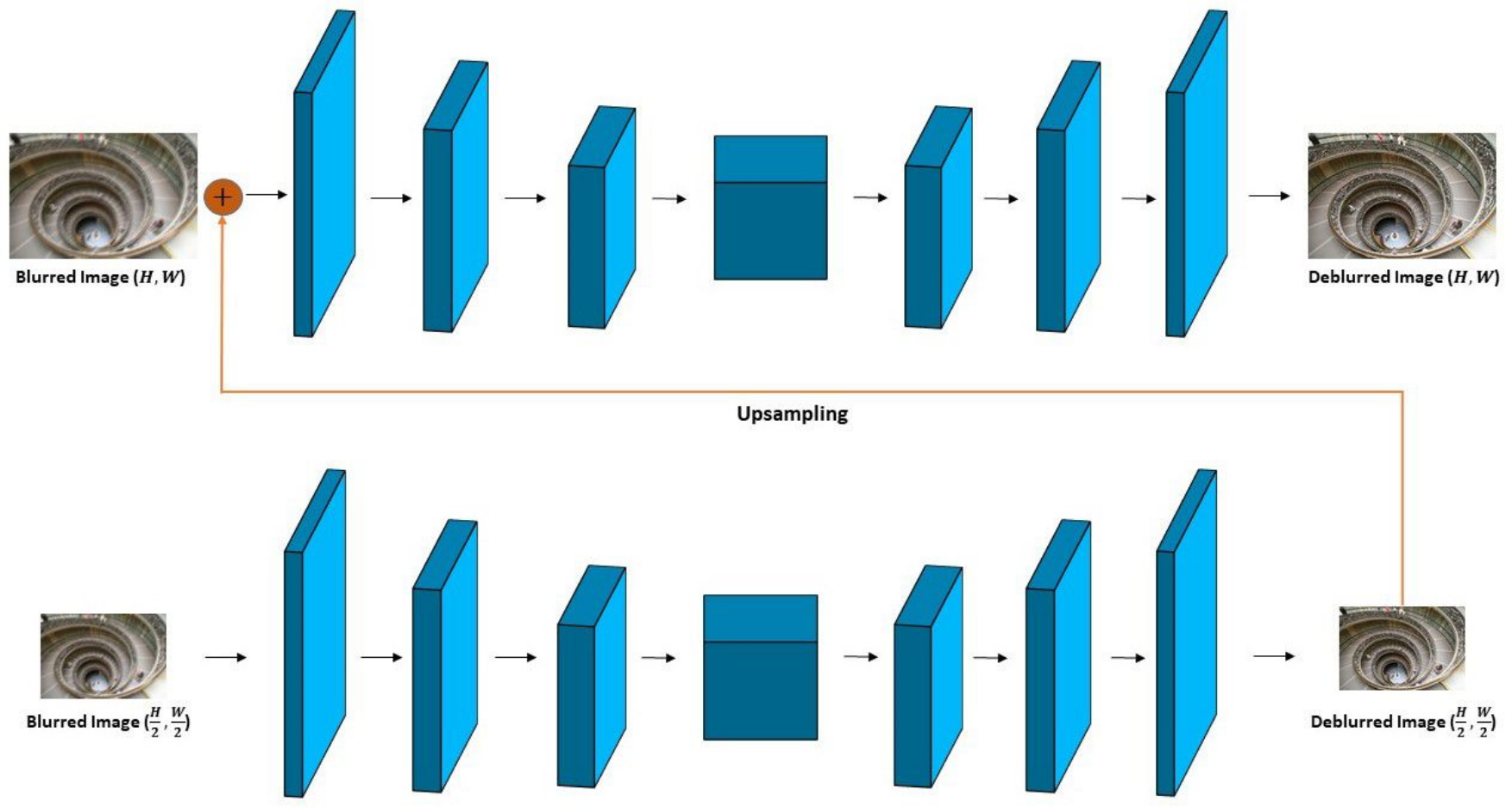}
    \caption{Multi-scale scheme}
    \label{fig:multScale}
\end{figure}

Multi-scale scheme progressively restores the latent image at different scales in a pyramidal manner. In other words, the algorithm is executed at the smallest scale of the image to estimate the coarsest outputs, e.g., blur kernel and restored image in blind deblurring, and then the resulting output image is combined with one at a finer scale to enhance the ultimate image recovery results \cite{cai2020dark}. This scheme has been used in various computer vision tasks, such as image segmentation \cite{long2015fully, eigen2015predicting}, image restoration \cite{nah2017deep, tao2018scale}, and video prediction \cite{mathieu2015deep}. This mechanism has successfully improved the quality of restored images in either prior-based optimization or deep neural image deblurring approaches \cite{tao2018scale}. An instance of multi-scale structure is presented in Fig.~\ref{fig:multScale}, in which $H$ and $W$ represent the height and width of the original input. 
\subsection{Attention}\label{sec:attention}
\par
Attention technique is motivated by some mechanisms of human perception \cite{itti1998model, woo2018cbam}. It is known that humans' visual system focuses on the salient features of subsequent frames of a scene rather than exploring the whole scene at once \cite{woo2018cbam}. Recently, this attention concept was introduced and integrated into a CNN structure to improve the network performance \cite{woo2018cbam, wang2017residual, hu2018squeeze}. There are two types of attention modules, namely, spatial attention and channel attention, each of which tries to extract different types of information from a whole image. 

\subsubsection{Spatial Attention Module}
\par
\begin{figure}[!h]
    \centering
    \includegraphics[width=0.9\textwidth]{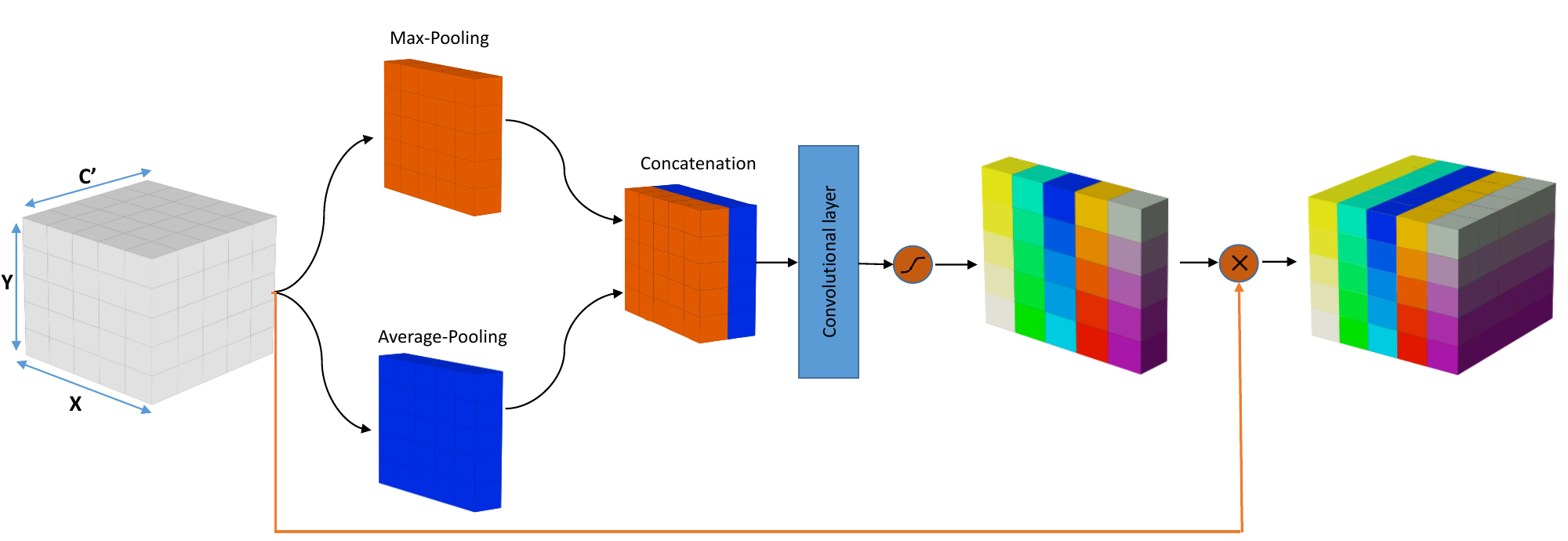}
    \caption{Spatial attention module scheme}
    \label{fig:spatial_attention}
\end{figure}

A spatial attention module extracts the spatial relationship between the features. That is, it specifies the location of useful information through all the channels. To clarify, let $\mathbf{Z} \in \mathcal{R}^{X \times Y \times C'} $ denotes the feature maps as shown in Figure.~\ref{fig:spatial_attention}.  $C'$ is the number of channels, and $(x,y)$ for $x\in X$ and $y \in Y$ are the spatial locations. Sub-sampling operations are used across all channels in every spatial location to compute a feature descriptor, and a 2D spatial attention map is generated by convolving the feature descriptor with a convolution layer. Then, the spatial attention map is assigned to all the channels. 
\citet{woo2018cbam} proposed convolutional block attention module (CBAM), leveraging the average-pooling as well as max-pooling operations, both being concatenated to create a feature descriptor as illustrated in Figure.~\ref{fig:spatial_attention}. 

\subsubsection{Channel Attention Module}
\par
\begin{figure}[!h]
    \centering
    \includegraphics[width=0.9\textwidth]{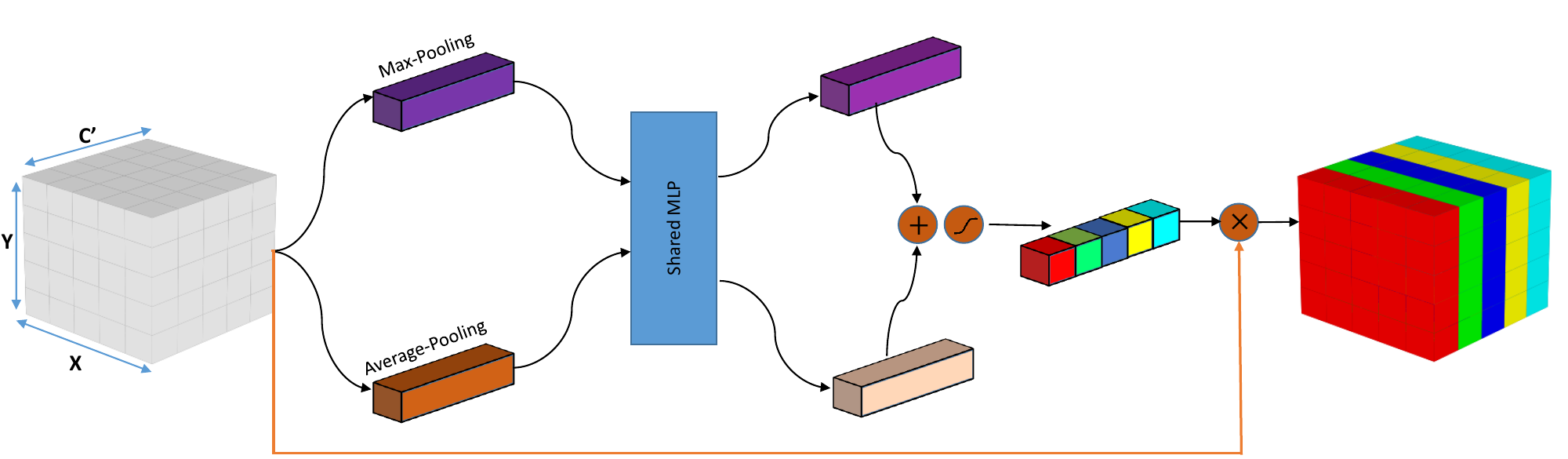}
    \caption{Channel attention module scheme}
    \label{fig:channel_attention}
\end{figure}

This attention map emphasizes the overall information available in each channel from all spatial locations. Similar to spatial attention module, it uses some sub-sampling operations, but they are applied within each channel to compute spatial statistics.  Although the average-pooling operation has been used frequently \cite{zhou2016learning, hu2018squeeze}, \citet{woo2018cbam} recently proposed the idea of applying both average-pooling and max-pooling operations to aggregate the spatial information since the latter operation can also provide proper information about the object features. Hence, these two spatial descriptors are generated and fed into a multi-layer perceptron (MLP) with a single fully-connected layer. Then, the sigmoid function is applied to the combination of descriptors' outputs to generate an attention map with $[0,1]$ values for each channel. Figure.~\ref{fig:channel_attention} illustrates the channel attention module of CBAM architecture \cite{woo2018cbam}. 


\section{Deep Neural Image Deblurring}
\label{sec:deep_deblurring}
\par
In this section, we briefly review a handful of works on non-blind deblurring. Then, we present a comprehensive review of deep neural architectures used in blind deblurring as there are a plenty of studies for this type of image deblurring.
\subsection{Non-blind Deblurring} 
\par
In the domain of non-blind deblurring, suppressing noise in the inversion process is the key to the recovery of a latent image with high-quality \cite{quan2021nonblind}. For that purpose, most approaches use either handcrafted or deep learning-based priors for an optimization procedure. In this section, we focus on the studies applying deep learning-based priors that are commonly incorporated into the optimization function in Eq.~\eqref{prior_obj}  as denoiser priors.
\par
\citet{schuler2013machine} use a multi-layer perception (MLP) network to deblur images in a non-blind manner. To suppress noise and eliminate artifacts, they additionally leverage a denoiser network post-processing the output of the image deblurring method.
Interestingly, \citet{zhang2017learning} propose a CNN-based denoiser network as a learned denoiser prior term for the prior-based optimization approach in Eq.~\eqref{prior_obj} to solve various image restoration problems, including image deblurring. They use the dilated convolution \cite{yu2015multi} to capture extensive receptive field and context information while regulating the network depth in the proposed denoiser. Their results show that learnable denoiser priors can outperform conventional statistical priors. In addition, \citet{xu2014deep} propose an image deblurring-based CNN that is constructed based on separable kernels extracted via singular value decomposition (SVD) \cite{perona1995deformable} and further decomposed into a small set of filters \cite{nah2017deep}. They also adopt a denoising CNN module \cite{eigen2013restoring} to remove the artifacts of a latent image. The entire network structure is developed and trained for uniform blur kernels, and for modeling more complex kernel structures, it is necessary to retrain the whole network.  To address the issue, \citet{ren2018deep} compute the low-rank approximations of separable blur kernels and incorporate them into their proposed generalized CNN network.   
\par
Concerning more sophisticated networks, \citet{kruse2017learning} propose an enhanced iterative fast Fourier transform (FFT) technique for non-blind image deblurring by leveraging CNNs in shrinkage fields \cite{schmidt2014shrinkage}. Shrinkage fields is a novel random field model that is built upon an enhanced form of half-quadratic optimization \cite{geman1995nonlinear} and is known to be effective for image restoration problems \cite{kruse2017learning}. Recently, \citet{gong2018learning} develop a recurrent gradient descent network (RGDN) as a learning optimizer which can learn an implicit prior for the optimization process and improve the performance. More elaborately, the proposed network would combine CNN with a gradient descent scheme in which the CNN elements of the gradient generator would tune the parameters.     

\subsection{Blind Deblurring}
\par
In a seminal work, \citet{hradivs2015convolutional} develop a CNN-based approach to deblur text documents in a blind manner. Their fundamental network structure is taken from the AlexNet \cite{krizhevsky2012imagenet}, with minor modifications in some hyperparameter settings, e.g., the number of layers and the number of filters. Following this study, \citet{sun2015learning} propose to use a CNN to predict the probabilities of motion blurs at each image patch. The output of this network is a set of blur candidates with various motion orientations and lengths forming the parameters of kernels. Given the likelihoods predicted by CNN, a Markov random field model is employed to combine all the patch-based blurs and build a dense non-uniform motion blur field. In addition, \citet{schuler2015learning} stack multiple convolutional layers to extract prominent features in a multi-scale fashion and mimic the conventional iterative optimization by estimating the kernel and the latent image alternatively.
Regarding blur kernel estimation, \citet{yan2016blind} propose a two-stage framework that concatenates a pre-trained deep network with a single regression network. The first network is trained to learn the feature maps of the blurred patches and classify them into three pre-defined blur types, including Gaussian blur, motion blur, and defocus blur.  Then, the next network would estimate the corresponding blur kernel parameters.
\par
\citet{chakrabarti2016neural}, as one of the well-known studies in this domain, designs and trains a multi-layer network to predict the frequency information (complex Fourier coefficients) of a deconvolution filter, which is applied to the input patch for the restoration process. Its primary goal is to estimate a single global blur kernel and subsequently restore the latent image in a non-blind fashion.  Image patch is also encoded into different frequency bands, including low-pass, band-pass, and high-pass, for its usage for varying sizes of image patch; this is called multi-resolution frequency decomposition. This encoding procedure would restrict the number of weights in the network alleviating the computational concern.

\begin{figure}[!t]
    \centering
    \includegraphics[width=0.9\textwidth]{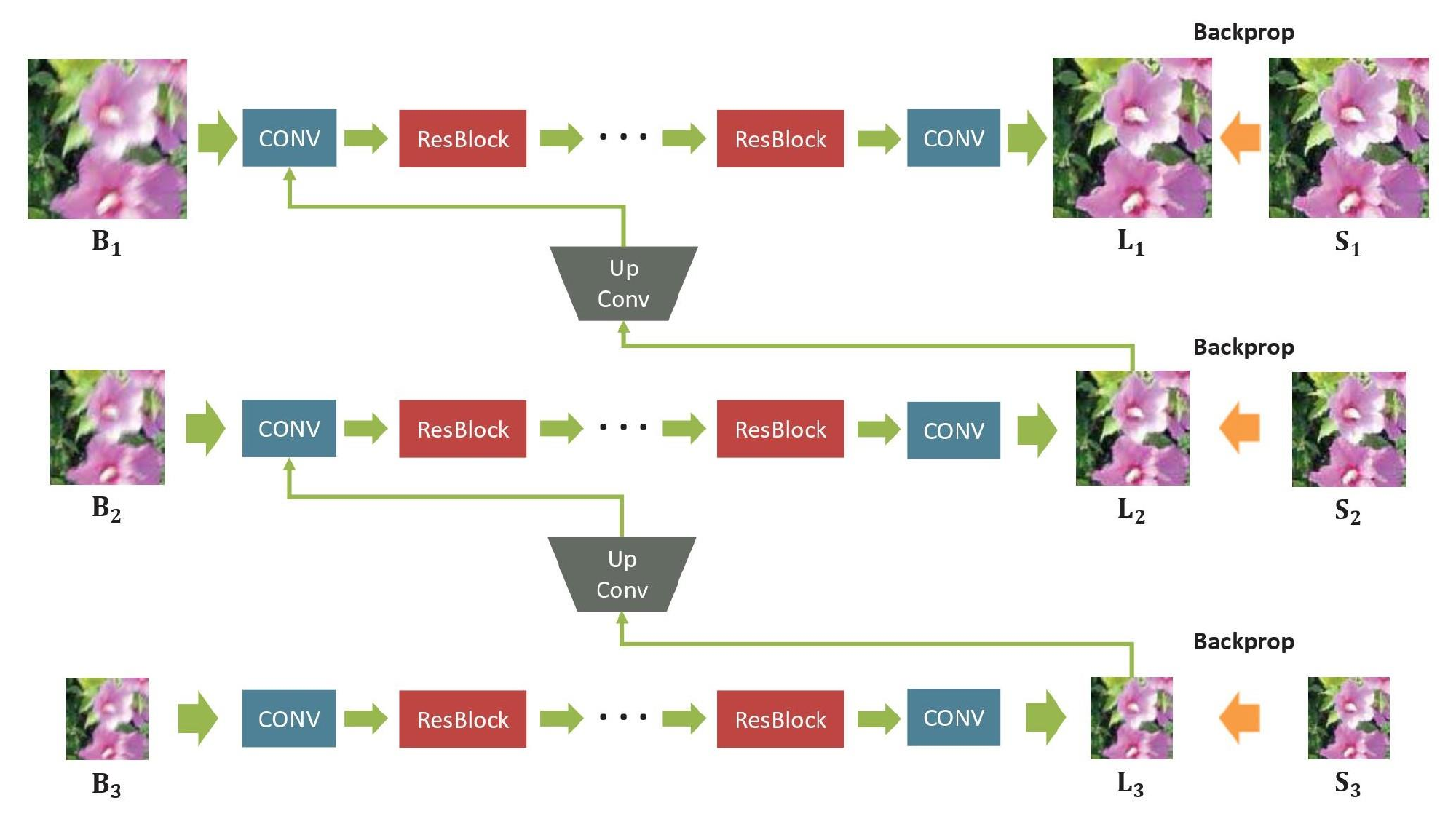}
    \caption{Multi-scale convolutional network scheme (copied from \cite{nah2017deep})}
    \label{fig:nah_multi}
\end{figure}

\par 
Among earlier works applying deep learning for blur kernel estimation, \citet{xu2017motion} apply a CNN-based structure to enhance and sharpen the edges of blurred images for better estimation of a blur kernel and thereby better restoration of a latent image. Their deep architecture consists of two sub-networks whose goals are to remove minor details and enhance the original structure of an image, respectively. Different from other earlier works, this edge sharpening algorithm is not paired with any heuristic approaches or multi-scale (coarse-to-fine) structure. Once the prominent edges are extracted and sharpened, the conventional alternating optimization method is leveraged to estimate the kernel and restore the latent image. Similarly, \citet{gong2017motion} develop a CNN-based model to learn and estimate the motion blur kernel for subsequent removal in the image deblurring process.
\par
Different from the studies discussed earlier incorporating neural structures for the estimation of a blur kernel, \citet{nah2017deep} introduce a multi-scale CNN architecture to present an end-to-end approach that directly restores latent images without any kernel estimation step. Instead, the deblurring procedure needs to acquire large enough receptive fields in order to handle very complicated blur kernels. One straightforward approach is to increase the number of convolutional layers, but this results in high computational cost in terms of training time. To impose a large receptive field, they consume three layers of deep networks in a coarse-to-fine manner, each of which possesses 19 residual blocks followed by a convolutional layer equating a feature map size with the dimension of ground truth images. Their residual blocks are a modified version of the original residual network \cite{he2016deep} in which batch normalization \cite{ioffe2015batch} as well as the ReLU (rectified linear unit) function after the shortcut connection are eliminated.  They show that this makes the algorithm converge faster while keeping the receptive field large enough by stacking several convolutional layers with residual blocks.  Interestingly, the ultimate latent image in a layer is concatenated with a finer layer's input to acquire as much structural information, such as the features of the coarser layer's outcome, as possible.  Although exploiting a multi-scale framework can improve the model's performance, it is computationally expensive due to the estimation procedures running over several scales \cite{ramakrishnan2017deep}. 
Figure \ref{fig:nah_multi} illustrates the multi-scale architecture. 

\begin{figure}[b!]
    \centering
    \includegraphics[width=0.9\textwidth]{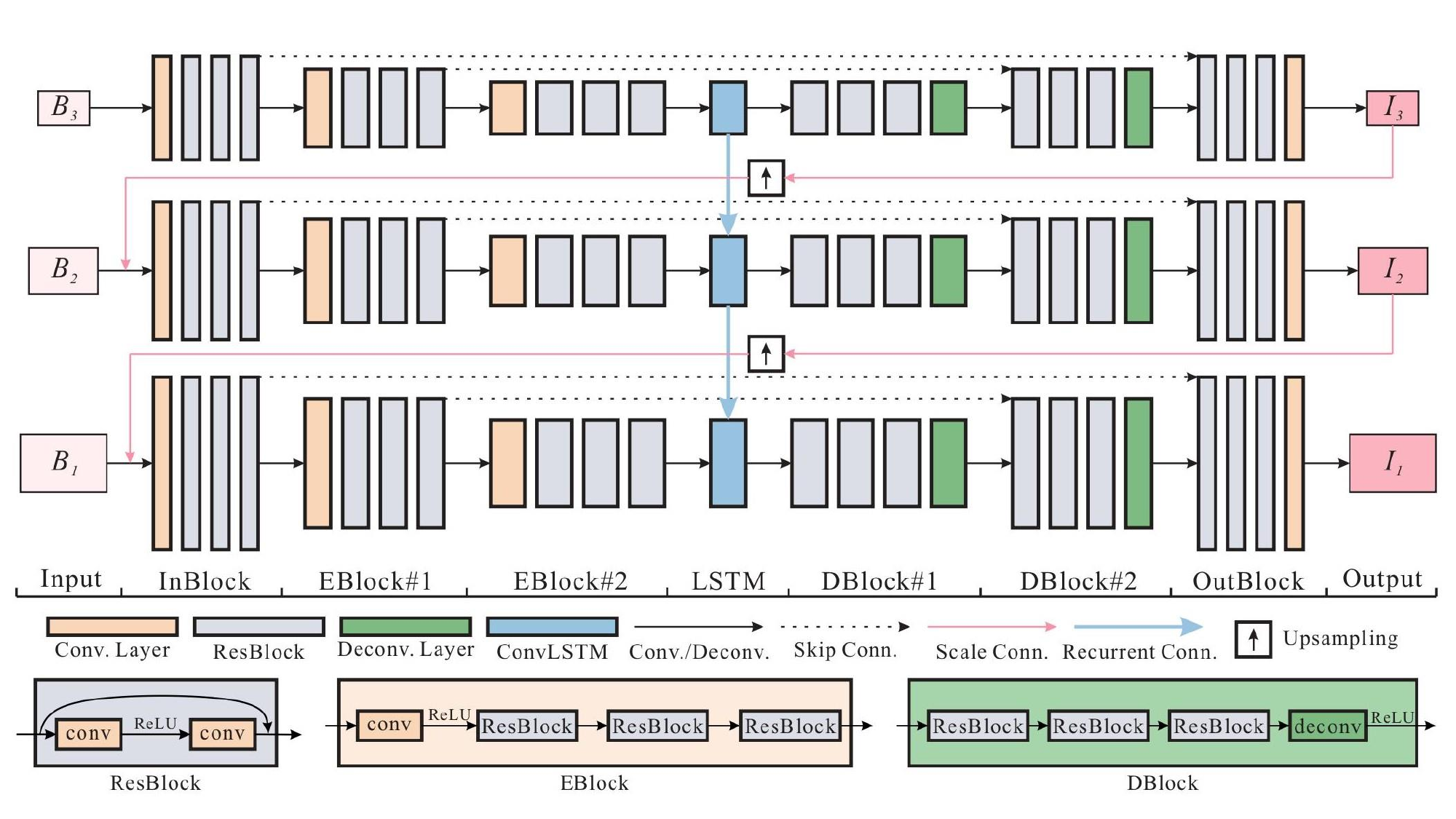}
    \caption{Scale-recurrent network architecture (copied from \cite{tao2018scale})}
    \label{fig:tao}
\end{figure}
\par
Inspired by the multi-scale mechanism and the residual blocks proposed in \citet{nah2017deep},  \citet{tao2018scale} propose a multi-scale recurrent encoder-decoder network in which ConvLSTM cells \cite{xingjian2015convolutional} are used as recurrent modules to integrate the information of a coarser latent representation layer into a finer scale as a hidden state, as depicted in Figure \ref{fig:tao}.  This approach aggregates the feature maps across all scales. The hidden state might transfer some  critical information about the intermediate latent image and blur kernel to the subsequent scale \cite{tao2018scale}. This approach shares the network weights across several scales to improve stability and diminish the number of trainable parameters causing expensive training costs elsewhere \cite{zhang2019deep}. 
\par
\begin{figure}[t]
    \centering
    \includegraphics[width=0.4\textwidth]{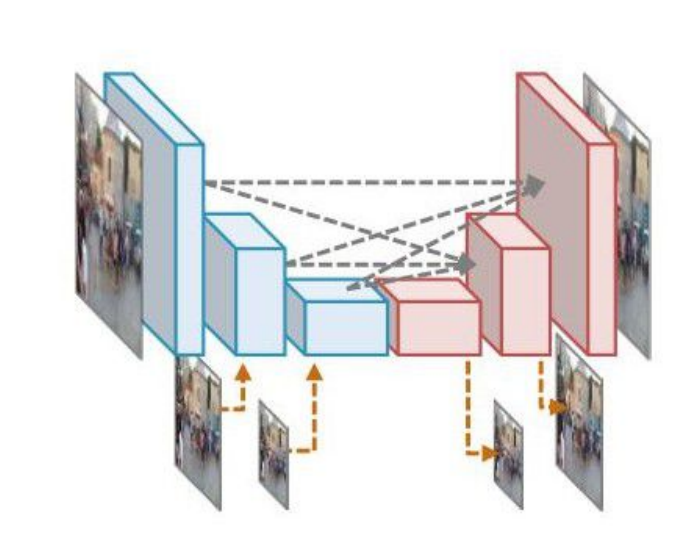}
    \caption{Multi-input multi-output U-Net (MIMO-UNet) architecture (copied from \cite{cho2021rethinking})}
    \label{fig:inputOutput}
\end{figure}

In addition to the multi-scale structure proposed by \citet{nah2017deep} and the scale-recurrent scheme introduced by \citet{tao2018scale} that are well-known architectures in this field, there are other works promoting structural advances in the literature. \citet{zhang2019deep} propose a deep multi-patch hierarchical deblurring network to improve deblurring results. They discuss that solely increasing the network depth in a simple multi-scale mechanism cannot improve the restoration results. Instead, they leverage spatial pyramid matching \cite{lazebnik2006beyond} that imposes a coarse-to-fine structure over multiple image patches in a hierarchical representation. Their quantitative outcomes show the superiority of the proposed structure with spatial pyramid matching to other state-of-the-art methods in both performance and runtime.  As the most recent study imposing coarse-to-fine structures, \citet{cho2021rethinking} propose a multi-input multi-output U-Net (MIMO-UNet). The encoder of their single U-Net structure takes multi-scale input images and integrates all the extracted features by using a newly developed asymmetric feature fusion module that uses convolutional layers to combine the multi-scale features. Then, the decoder returns multi-scale output images that are used to train the network in the coarse-to-fine structure. Figure \ref{fig:inputOutput} displays their proposed deep architecture. 
\par
Instead of applying independent weights as in \citet{nah2017deep} or sharing weights across various scales as in \citet{tao2018scale}, \citet{gao2019dynamic} propose parameter selective sharing in an encoder-decoder structure with nested skip connections to capture more constructive features. They argue that both independent weights and  shared weights are not effective in integrating weights of different scales. As such, they introduce a parameter selective sharing technique.  This technique leverages independent weights for feature extraction modules in each scale, but it assigns the same weights across all scales for nonlinear transformation modules. They also adopt nested skip connections, similar to DenseNet \cite{huang2017densely}, with a minor change in the number of links at the last convolution layer and the operator for fusing features \cite{gao2019dynamic}. Regarding the weight sharing scheme, \citet{zhang2018dynamic} propose a spatially variant recurrent neural network where the weights are learned by a separate CNN.  Their proposed network has a large receptive field showing promising performance for the restoration of latent images. 

\par
\begin{figure}[b!]
    \centering
    \includegraphics[width=0.9\textwidth]{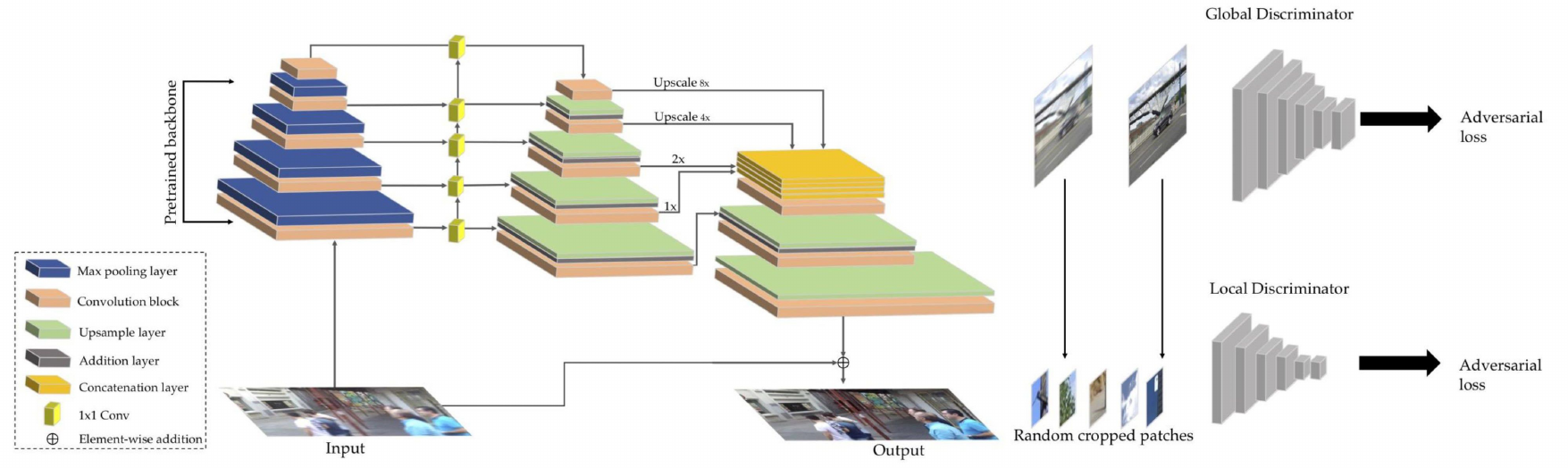}
    \caption{DeblurGAN-v2 architecture (copied from \cite{kupyn2019deblurgan})}
    \label{fig:deblurGAN2}
\end{figure}

On the other hand, generative networks have been frequently used to enhance networks performance and reduce required computations. \citet{ramakrishnan2017deep} propose a densely connected generative network where the discriminator is a Markovian patch discriminator \cite{li2016precomputed} that operates convolutional layers on local patches rather than a full image to distinguish the local patch textures. \citet{nimisha2017blur} propose a GAN structure combined with an encoder-decoder network to restore a latent image while using the extracted features of the encoder segment for the input of the GAN structure \cite{nimisha2017blur}. This approach can diminish the computational cost for model training and can handle both uniform and non-uniform blurs. Among all studies leveraging GAN structures for image deblurring, DeblurGAN \cite{kupyn2018deblurgan} is the most well-known structure.  It is established based on conditional GAN \cite{mirza2014conditional}, and its critic (discriminator) network is a Wasserstein GAN \cite{arjovsky2017wasserstein} with gradient penalty improvement \cite{gulrajani2017improved}. 
This architecture is further enhanced by adding the feature pyramid network structure \cite{lin2017feature} and using a double-scale discriminator for both local (patch-based) \cite{isola2017image} and global (full-image) features; this new approach is called DeblurGAN-v2 \cite{kupyn2019deblurgan}. It is shown that DeblurGAN-v2 has less runtime and competitive performance compared to the former version of DeblurGAN. \citet{kupyn2019deblurgan} consider several feature extractor backbones, including Inception-ResNet-2 \cite{szegedy2017inception}, MobileNet \cite{sandler2018mobilenetv2}, and MobileNet with depthwise separable convolutions \cite{chollet2017xception} to evaluate their performance and efficiency and select the best feature extractor architecture. Figure \ref{fig:deblurGAN2} illustrates the DeblurGAN-v2 architecture.   

\par
Instead of relying on a single GAN structure, \citet{zhang2020deblurring} propose a fusion of two GAN structures for both blurring and deblurring, referred to as blurring GAN and deblurring GAN. The generator of the blurring GAN tries to generate blurred images out of real sharp images, and the discriminator compares the generated blurred images with other actual blurred images to fool the generator network and return more realistic blurred images. In the deblurring GAN network, multiple pairs of an original sharp image and the corresponding blurred image generated by the blurring GAN are used to learn the deblurring process in a GAN structure. By construction, this network learns how to generate realistic blurring effects and how to recover latent images by using ground truth images. More recently, \citet{zhao2021gradient} introduce a conditional GAN with dense blocks \cite{huang2017densely} to improve the feature extraction process in the generator network by fusing different kinds of features and using the resulting outcome as the output of the block. They also adopt instance normalization \cite{ulyanov2016instance} rather than batch normalization so that the normalization applies to each sample data avoiding instance-specific mean and covariance shift and hence becomes more suitable for the image generation task. Meanwhile, their discriminator is built based on PatchGAN \cite{isola2017image} that discriminates real images from fake ones at the scale of patches and models high-frequency structures with fewer parameters relative to general GAN structures learning from the whole size of images. To improve the training process of the network, they employ the gradient loss in addition to common loss functions.
\par
The attention mechanism discussed in Section \ref{sec:attention} is introduced recently in the image deblurring field and becomes popular as it is capable of extracting blur characteristics along with their corresponding locations.  \citet{purohit2020region} introduce self-attention \cite{zhang2019self} and dense deformable modules in their encoder-decoder structure to effectively learn the global and local spatial transformation and characterize the non-uniform blurs. These modules can potentially identify spatially varying blurs and the spatial relationships of the underlying features. However, incorporating these modules requires more computational capacity \cite{chen2021attention}.  \citet{xu2021attentive} propose integrating attention modules, including both spatial and channel attention, into a multi-scale encoder-decoder architecture to handle blurs with large spatial variations and generalize the network for the usage in various types of non-uniform blurred images. The incorporated spatial and channel attention modules in both encoder and decoder structures extract features that are more responsible for the blurring effect and effectively retrieve spatially-varying image regions.
The channel attention module can also help improve the generalizability of the CNN in the deblurring process \cite{xu2021attentive}.
 \par
As an extension, \citet{chen2021attention} integrate two modules, adaptive-attention and deformable convolution \cite{dai2017deformable, zhu2019deformable}, into a vanilla CNN to improve the quality of restored images. The former module adaptively determines in which way the spatial and channel attention modules should be combined for an optimal arrangement, either sequentially or in parallel, by employing auxiliary classifiers. The deformable convolutional module can handle various geometric structures in different spatial regions that are commonly observed in dynamic scenes. The integration of these modules in a CNN structure effectively captures image features and better restores latent images according to qualitative evaluations and quantitative metrics, including peak signal-to-noise ratio (PSNR) and structural similarity measure (SSIM). There are other studies integrating attention-based modules to extract constructive features in the literature. \citet{li2021single} propose cross-layer feature fusion and consecutive attention modules which are incorporated into the generator of a GAN structure. The cross-layer feature fusion module integrates the outputs of the last three encoder layers rather than those of the last single encoder (common arrangement in the literature) to obtain the most original features and improve the resolution of feature map. To retrieve the most correlated textures of an image, a consecutive attention module is added on top of the last decoder layer as well. This consecutive attention module is basically the criss-cross attention module \cite{huang2019ccnet} that has two subsequent attention blocks to capture the full-image dependencies and contextual information within criss-cross path.    
\par
\citet{tsai2021banet} develop a blur-aware attention module, constructed by multi-kernel strip pooling \cite{hou2020strip} and attention refinement parts, to capture global and local information of blur effects. The blur-aware attention module requires less memory and computational resources compared to the self-attention module \cite{zhang2019self, purohit2020region, tsai2021banet}.
While leveraging attention modules, \citet{luo2021bi} propose to configure two distinct branches for capturing both RGB content features and motion-related spatiotemporal features. The two types of extracted features are integrated across their proposed nonlocal fusion layer that performs  the double attention operation \cite{chen20182} to combine heterogeneous transformations in the encoder. They show this proposed network can enhance deblurring performance and restore high-quality images while remarkably alleviating the computational issues.

\begin{figure}[!b]
    \centering
    \includegraphics[width=0.6\textwidth]{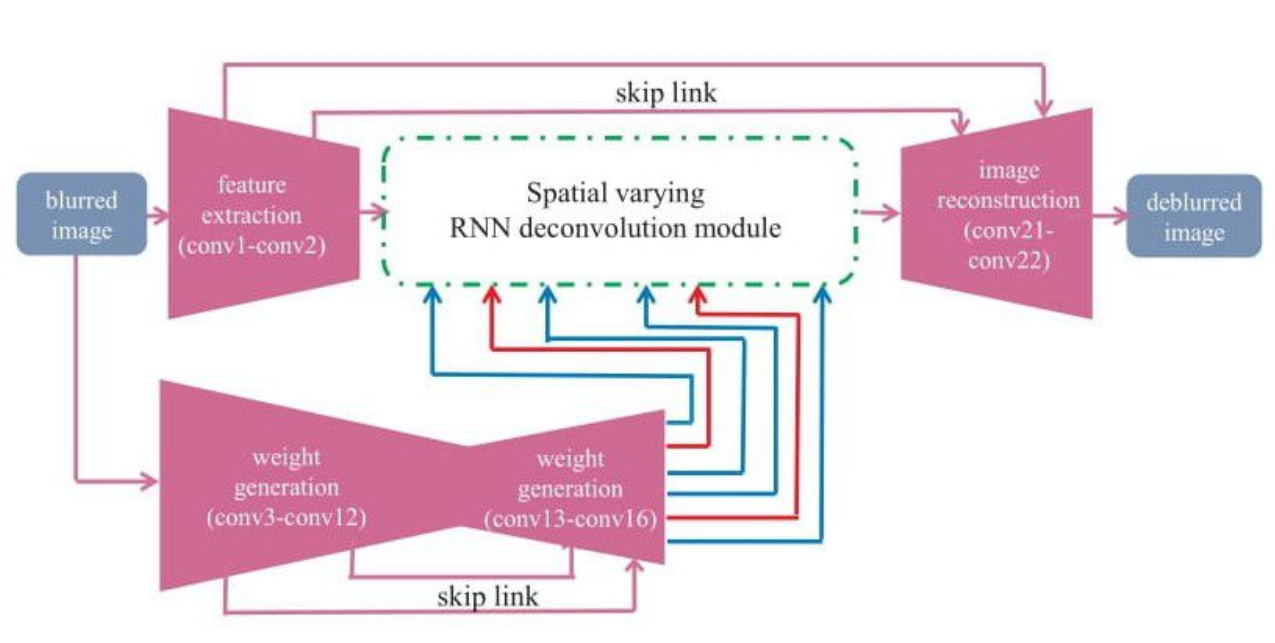}
    \caption{Spatially varying recurrent neural network (copied from \cite{ren2021deblurring})}
    \label{fig:li_spatial}
\end{figure}

\par
For more complicated deep neural structures, \citet{ren2021deblurring} propose a spatially varying RNN by using recurrent and convolutional layers. The network consists of a CNN-based feature extraction module, an RNN-based deblurring module, a CNN for generating the weights of the RNN, and an image reconstruction part. They use either one-dimensional (1D) or two-dimensional (2D) RNNs in their deblurring module. They conclude that, compared to the 1D RNN, the 2D RNN can learn more information in the same receptive field since it covers more spatial propagation through a three-way connection. The three-way connection of the 2D RNN enables to expand a region into a triangle 2D plane at each direction. Figure \ref{fig:li_spatial} illustrates the spatially varying RNN. As displayed, two CNNs are leveraged to extract the features and estimate the final deblurred image in the image reconstruction module. The spatial varying RNN would eliminate the blur, where its weights are generated by another CNN.

\begin{table}[h]
\begin{center} 
\caption{Structural specifications of blind deblurring papers in chronological order}
{
\resizebox{\textwidth}{!}{
\scriptsize
\begin{tabular}{lcccccccccc}
\hline
  & \multicolumn{6}{c}{Deep neural structures} & Multi Scale-approach & Skip connection & Attention module\\
 & Fully-Connected & \multicolumn{1}{c}{Convolutional} & \multicolumn{1}{c}{Encoder-Decoder} & LSTM/Recurrent & GAN & Residual \\
\hline
\citet{hradivs2015convolutional} & & \checkmark  & &   &  &  & & &\\
\citet{sun2015learning} & \checkmark & \checkmark  & &   &  &  & & & \\
\citet{schuler2015learning} & & \checkmark  & &   &  &  & \checkmark &  & \\
\citet{yan2016blind} & \checkmark & \\
\citet{chakrabarti2016neural} & \checkmark & \checkmark  & &   &  &  & &  \\
\citet{gong2017motion} & & \checkmark & \checkmark &    &  & & & \checkmark \\
\citet{ramakrishnan2017deep} & & \checkmark & & & \checkmark & & & \checkmark\\
\citet{nah2017deep} & & \checkmark & &   &  & \checkmark & \checkmark &  &\\
\citet{xu2017motion} & & \checkmark & &   &  & & &  \\
\citet{nimisha2017blur} & & \checkmark & \checkmark &   & \checkmark & \checkmark & & \\
\citet{li2018learning}& & \checkmark & &   &  & & \\
\citet{kupyn2018deblurgan} & & \checkmark & \checkmark & & \checkmark & \checkmark & & \checkmark\\
\citet{zhang2018dynamic} & & \checkmark & \checkmark & \checkmark & & & & \checkmark\\
\citet{tao2018scale}& & \checkmark & \checkmark & \checkmark  &  & \checkmark & \checkmark & \checkmark\\
\citet{gao2018stacked} & & \checkmark & \checkmark \\
\citet{shen2018deep} & & \checkmark &  & & & \checkmark & \checkmark\\
\citet{zhang2019deep} & & \checkmark & \checkmark &   &  & & \\
\citet{aljadaany2019douglas} & &  \checkmark & &   &  & & \\
\citet{kupyn2019deblurgan} & &  \checkmark & \checkmark &   & \checkmark & & & \checkmark\\
\citet{gao2019dynamic} & & \checkmark & \checkmark & & & & \checkmark & \checkmark\\
\citet{li2019deep}& & \checkmark & &   &  & & \\
\citet{chen2019u} &  & \checkmark & \checkmark &   &  & & & \checkmark & \\
\citet{shen2019human} & & \checkmark & \checkmark & & & &\checkmark&& \checkmark \\
\citet{cai2020dark} & & \checkmark & \checkmark &   &  & \checkmark & \checkmark & \checkmark &\\
\citet{purohit2020region} & & \checkmark  & \checkmark &   &  & & & \checkmark& \checkmark\\
\citet{lin2020learning} & & \checkmark & \checkmark &   & \checkmark & \checkmark & & \checkmark \\
\citet{chen2020blind} & & \checkmark & \checkmark &   &  & & & \checkmark\\
\citet{zhang2020deblurring} & \checkmark & \checkmark & & & \checkmark & \checkmark  \\
\citet{ren2020neural} & \checkmark & \checkmark & \checkmark & & \checkmark & & & \checkmark & \\
\citet{asim2020blind} & & \checkmark & \checkmark & & \checkmark\\
\citet{xu2021attentive} & & \checkmark & \checkmark & & & \checkmark & \checkmark & \checkmark & \checkmark  \\
\citet{zhao2021gradient} & & \checkmark & \checkmark & & \checkmark & & & \checkmark &  \\
\citet{chen2021attention} & \checkmark & \checkmark & \checkmark & & & \checkmark & & \checkmark & \checkmark\\
\citet{cho2021rethinking} & & \checkmark & \checkmark & & & \checkmark & \checkmark & \checkmark & \checkmark\\
\citet{dong2021deep} & & \checkmark & & & & \checkmark & \checkmark &  &\\
\citet{luo2021bi} & & \checkmark & \checkmark &  & & \checkmark & & \checkmark & \checkmark\\
\citet{li2021single} & & \checkmark & \checkmark & & \checkmark & \checkmark & & \checkmark & \checkmark\\
\citet{ren2021deblurring} & & \checkmark & \checkmark & \checkmark & & & & \checkmark & \\
\citet{wu2021stack} & & \checkmark & \checkmark & \checkmark & & \checkmark & \checkmark &  \\
\citet{hu2021pyramid} & & \checkmark & \checkmark &  & & & \checkmark\\
\citet{quan2021gaussian} & & \checkmark & \checkmark  & \checkmark & & & \checkmark & \checkmark & \checkmark \\
\citet{tsai2021banet} & & \checkmark & \checkmark & & & & & \checkmark & \checkmark\\
\\
\hline
\label{specification}
\end{tabular} }}
\end{center}
\end{table}

\par
There are some studies relating neural image deblurring to the conventional prior-based optimization approach shown in Eq.~\eqref{prior_obj}. 
\citet{aljadaany2019douglas} develop two deep learning-based proximal operators associated with the data fidelity and prior terms in Eq.~\eqref{prior_obj} and solve the prior-based optimization problem by using Douglas-Rachford iterations \cite{eckstein1992douglas}. The proposed proximal operators are modeled by CNNs to estimate both terms in Eq.~\eqref{prior_obj}. More interestingly, \citet{cai2020dark} embed image priors of dark channel \cite{he2010single} and bright channel \cite{yan2017image} into a CNN structure. The feature outputs of dark channel and bright channel layers are concatenated with the original feature map in encoder and decoder components in order to extract and recover the prior knowledge for those channels from blurred images. To enforce sparsity on the feature maps, the training procedure applies $L1$-regularization. 
They also introduce image full-scale exploitation (IFSE), a multi-scale structure that leverages both fine-to-coarse and coarse-to-fine directional schemes to acquire all the information flows across the scales.  Their reported results demonstrate promising performance in comparison to other multi-scale structures \cite{nah2017deep, tao2018scale}, and they conclude that the embedded layers associated with dark channel and bright channel effectively improve the quality of restored images. 
Meanwhile, \citet{li2019deep} introduce an algorithm unrolling technique to make a connection between the prior-based optimization and deep neural image deblurring for better interpretability of a model.  Specifically, they unroll the conventional total-variation (TV) regularization algorithm to build a deep neural network for image deblurring.    

\par
While most studies in the literature share similar techniques and network structures, there are some others with unique structures and implementations.  \citet{hu2021pyramid} develop a multi-scale pyramid neural architecture search approach (PyNAS) to optimize architecture designing hyperparameters associated with patches, scales, and cell operators to efficiently handle the non-uniform blurs in dynamic scene deblurring problems.  The optimization process involves gradient-based search and their proposed hierarchical search strategies for automatic hyperparameter learning. On the other hand, \citet{quan2021gaussian} introduce a Gaussian kernel mixture network to alleviate spatially variant defocus blur. Their network adopts a scale-recurrent attention module that incorporates Conv-LSTM elements into the attentive encoder-decoder backbone \cite{mao2016image}. This network also includes a Gaussian convolution module, as a part of feature extractor, and it is built based on a set of pre-defined 2D Gaussian kernel convolutional layers to apply to each color channel of the blurred image. 

\par
Tables~\ref{specification} and~\ref{tab:contr}, respectively, summarize structural specifications and main contributions of all deep neural image deblurring studies reviewed in this paper. The provided tables show that the majority of deep neural image deblurring structures consist of convolutional layers and the encoder-decoder architecture. In addition, skip connection is widely used when developing deep networks while attention module is a more recent development adopted in this domain.

\begin{table}[!t]
\begin{center} 
\caption{Specific contributions of blind deblurring papers in chronological order}
{
\resizebox{\textwidth}{!}{
\begin{tabular}{ll}
\hline
Studies  & Contributions\\
\hline
\citet{hradivs2015convolutional} & CNN for text documents\\
\citet{sun2015learning} &  CNN for predicting the oriented motion vectors and length\\
\citet{schuler2015learning} & CNN for extracting constructive features for further iterative optimization \\
\citet{yan2016blind} & Classifying the three pre-defined blur types and estimating blur kernel parameters\\
\citet{chakrabarti2016neural} & Estimating the global blur kernel by predicting the freuqency information of the deconvolution filter \\
\citet{gong2017motion} & CNN for estimating the motion flow model\\
\citet{ramakrishnan2017deep} & Proposing densely connected generative network with dilated convolution in generator and Markovian patch discriminator. \\
\citet{nah2017deep} & Directly restoring the latent image by proposing multi-scale CNN structure and residual blocks\\
\citet{xu2017motion} & CNN for sharpening the edges of blurred image for further iterative optimization \\
\citet{nimisha2017blur} & Combination of encoder-decoder network with GAN to generate blur-invariant features\\
\citet{li2018learning} & Proposing a data-driven discriminative prior using CNN  \\
\citet{kupyn2018deblurgan} & Wasserstein GAN and Conditional GAN for image deblurring\\
\citet{zhang2018dynamic} & Proposing a spatially variant recurrent neural network that its weights are trained by a deep CNN\\
\citet{tao2018scale} &  A multi-scale recurrent network with shared weights in scales and adopting ConvLSTM cells\\
\citet{gao2018stacked} & A convolutional auto-encoder (CAE) for spatial targets images\\
\citet{shen2018deep} & Incorporating global semantic prior into the multi-scale CNN with residual blocks for blurred face images\\
\citet{zhang2019deep} & A deep multi-patch hierarchical network by employing spatial pyramid matching approach\\
\citet{aljadaany2019douglas} & Developing two proximal operators for data fidelity and prior terms using CNN\\
\citet{kupyn2019deblurgan} &  Enhancing the DeblurGAN \cite{kupyn2018deblurgan} by introducing feature pyramid structure and double-scale discriminator \\
\citet{gao2019dynamic} & Proposing nested skip connections and parameter selective sharing for encoder-decoder network\\
\citet{li2019deep} & Adopting algorithm unrolling technique to connect neural networks with the conventional iterative algorithms\\
\citet{chen2019u} & Introducing a deep-stacked of a convolutional auto-encoder with U-Net structure for spatial targets images\\
\citet{shen2019human} & Incorporating a supervised attention mechanism into a multi-branch deblurring model\\
\citet{cai2020dark} &  Proposing a Dark and Bright Channel Priors Embedded Network  with image full scale exploitation structure \\
\citet{purohit2020region} & Introducing self-attention and dense deformable modules into the encoder-decoder structure\\
\citet{lin2020learning} & A generator network to learn the face sketches for blurred face images\\
\citet{chen2020blind} &  Proposing the deblurring noise suppression block in the U-Net structure\\
\citet{zhang2020deblurring} & Fusion of two GAN structures for both blurring and deblurring process.\\
\citet{ren2020neural} & Proposing a joint deep image prior for blur kernel and latent image estimation using autoencoder and fully-connected structures\\
\citet{asim2020blind} & Proposing deep generative network for estimation of blur kernel and latent image using generative networks\\
\citet{xu2021attentive} &  Introducing spatial and channel  attention modules into encoder-decoder network for image deblurring\\ 
\citet{zhao2021gradient} & A Conditional GAN structure with dense blocks\\
\citet{chen2021attention} & Integrating adaptive-attention and deformable convolution modules with CNN\\
\citet{cho2021rethinking} & Proposing a novel coarse-to-fine structure as multi-input multi-output U-Net (MIMO-UNet)\\
\citet{dong2021deep} & A CNN architecture to detect outliers and alleviate their impact on the deblurring process\\
\citet{luo2021bi} & A bi-branch structure for heterogeneous transformations on motion and RGB content features \\
\citet{li2021single} & Incorporating cross-layer feature fusion and consecutive attention modules into the GAN structure\\
\citet{ren2021deblurring} & Proposing a spatially varying RNN, whose weights are generated by a CNN structure\\
\citet{wu2021stack} & Stacking two scale-recurrent networks for blurred face images\\
\citet{hu2021pyramid} & Developing a novel hierarchical multi-scale neural search approach\\
\citet{quan2021gaussian} & Proposing a Gaussian kernel mixture network with scale-recurrent attention module\\
\citet{tsai2021banet} & Proposing blur-aware attention network for blind image deblurring\\
\hline
\label{tab:contr}
\end{tabular} }}
\end{center}
\end{table}

\subsection{Deep Learning-based Image Priors}
\par
\begin{figure}[!h]
    \centering
    \includegraphics[width=0.95\textwidth]{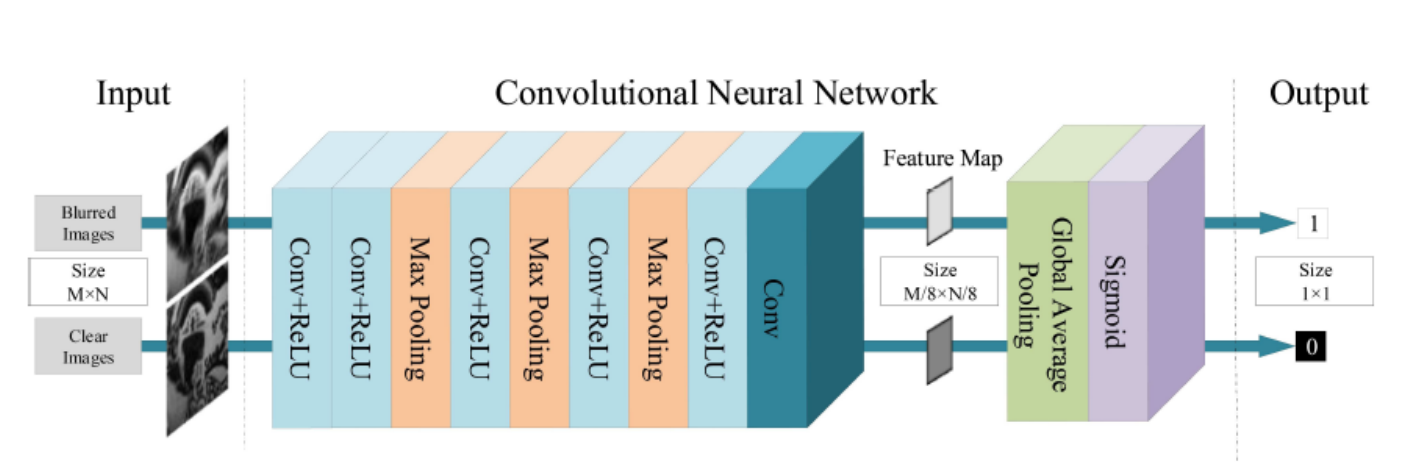}
    \caption{Data-driven discriminative prior architecture (copied from \cite{li2018learning})}
    \label{fig:li_discr}
\end{figure}

In the literature, there are some works applying deep learning techniques to extract the inherent information of images for its further usage in a conventional deblurring task. \citet{dong2021deep} propose a deep outlier detection technique using a deep CNN. Their algorithm estimates a confidence map of a blurred image to assign weights to each pixel that indicates the degree of being an outlier. These outlier weights are attached to the data fidelity term in Eq.~\eqref{prior_obj} making it as a weighted loss, and this restricts the impact of outlier pixels on the image deblurring procedure. \citet{li2018learning} propose a data-driven discriminative prior that leverages binary classifications from a deep CNN shown in Figure~\ref{fig:li_discr}. They believe that an image prior should be compatible more with clear images than with degraded ones to restore a favorable latent image.  In this regard, they design a network producing binary outputs where zero and one refer to a clear image and a blurred image, respectively, and use this information as an image prior ($P(I)$) in Eq.~\eqref{prior_obj}. The network consists of multiple stacked convolutional layers and is constructed by using a multi-scale training approach that randomly modifies the size of input images for robustness purpose \cite{li2018learning}.  To make the network flexible with varying sizes of inputs in terms of widths and heights, they use a global average pooling layer \cite{lin2013network} instead of a fully connected layer, and this allows converting a feature map of any size into a scalar value. 

\par
With significant importance in deep learning-based prior development, \citet{ulyanov2018deep} introduce a deep image prior that does not require pre-training of a model from a large set of images. In this approach, image priors are obtained by fitting a generator network to a single degraded image rather than learning network parameters from a large set of blurred images. 
The U-Net architecture \cite{ronneberger2015u} is adopted for image generation while learning a mapping function of $\hat{I}=f_{\theta}(z)$ as an encoder/decoder network, where $z$, $\theta$, and $\hat{I}$ are random samples, random parameters of the network, and the restored image, respectively. The restored image can be generated by optimizing the network parameters ($\theta$) and capturing the image statistics of a single blurred image. Although this prior is developed for an image deblurring task, it can be applied to other image restoration tasks, such as super-resolution and image inpainting \cite{ulyanov2018deep}.
Inspired by this study, \citet{cheng2019bayesian} investigate a Bayesian approach for the deep image prior. They discuss that the deep image prior can be interpreted as a stationary zero-mean Gaussian process since the number of channels in every layer goes to infinity. With this Bayesian architecture, posterior inference can be made for the deep image prior.


\begin{figure}[!b]
    \centering
    \includegraphics[width=0.95\textwidth]{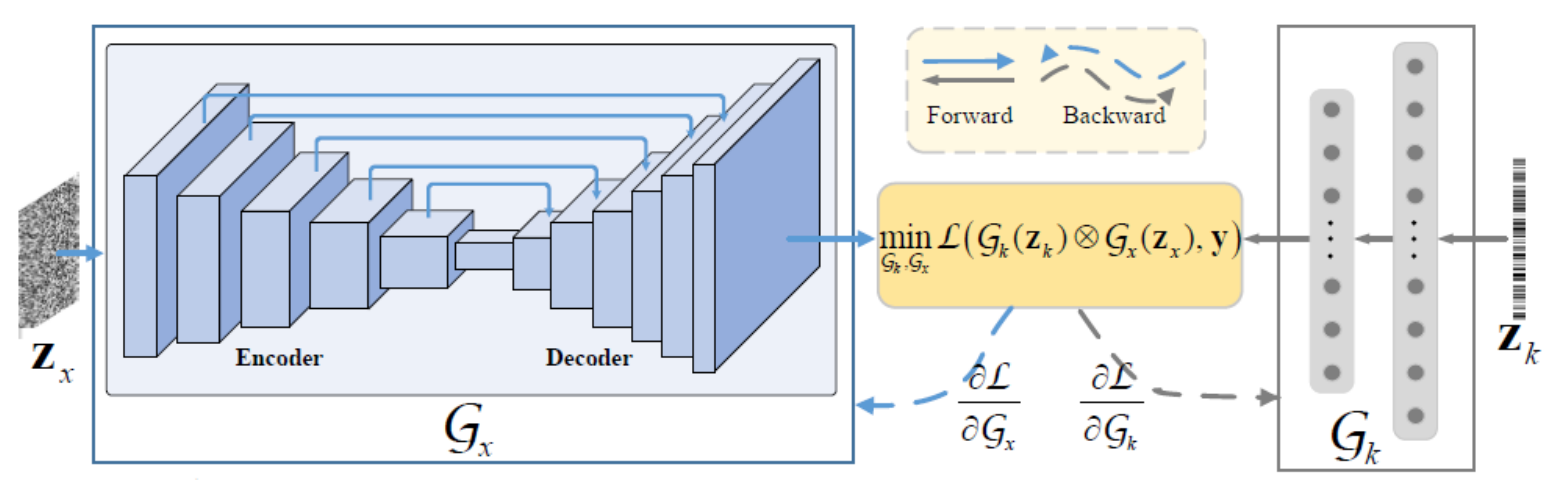}
    \caption{Deep image prior for blind deblurring (copied from \cite{ren2020neural}); an encoder-decoder network on the left and a fully connected network on the right.}
    \label{fig:ren_dip}
\end{figure}
\par
\begin{figure}[!h]
    \centering
    \includegraphics[width=0.95\textwidth]{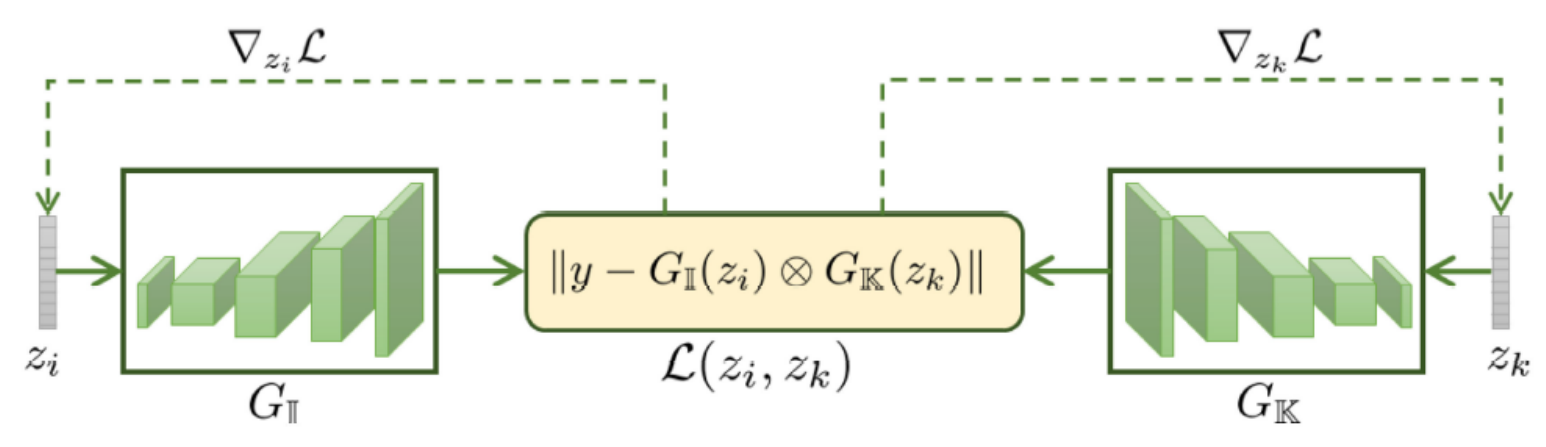}
    \caption{Deep generative prior for blind deblurring (copied from \cite{asim2020blind})}
    \label{fig:asim_gdip}
\end{figure}

More recently, \citet{ren2020neural} propose a joint deep image prior structure that applies to both kernel and latent image.  As shown in Figure \ref{fig:ren_dip}, an encoder-decoder network (deep image prior network) and a fully connected network are used to obtain the deep priors and estimate the latent image and kernel, respectively.    
Meanwhile, \citet{asim2020blind} introduce priors using deep generative network that consists of a pre-trained GAN ($G_\mathbf{I}$) and a VAE ($G_\mathbf{K}$) for estimating the latent image and blur kernel, respectively; see Figure~\ref{fig:asim_gdip}.  Different from other deep image priors, this generative prior needs to be trained on a large dataset whereas others require only a single blurry image to extract the statistics of the image. 

\subsection{Specific Applications: Face and Remote Sensing Images} 
\par
Although deblurring techniques can be used for any types of images, there are some special applications where particular deep learning architectures can be very useful.  The most practical applications include face and space target deblurring. In the face deblurring application, face images typically share similar semantic information as they involve the same major objects, e.g., eyes and nose. As such, a network structure emphasizes more on capturing semantic information for a deblurring task. \citet{shen2018deep} propose a multi-scale CNN with residual blocks that is similar to the multi-scale network in \citet{nah2017deep} but with primary modifications in network structures. They use two scales and contruct their network with fewer ResBlocks. Additionally, they incorporate  global semantic prior which is the probability maps of semantic labels and can be extracted by using the face parsing network \cite{liu2015multi}. \citet{lin2020learning} develop a generator network to learn and estimate the face sketches out of blurred face images. Then, the estimated sketch of the blurred image is used to estimate the blur kernel and restore the latent image in a conventional optimization-based manner. In a recent work, \citet{wu2021stack} propose to stack two scale-recurrent networks for recovering blurred face images as implemented in \citet{tao2018scale}. They point out that the stacking strategy efficiently increases the network depth so is better than increasing the number of convolutional layers which can inflate model complexity significantly. 
\par
Meanwhile, deep neural image deblurring has been frequently used for space target images, generally obtained by remote sensing. \citet{gao2018stacked} propose a convolutional auto-encoder (CAE) architecture for the usage in this specific type of images. Their network consists of convolutional and deconvolutional layers to extract features for the deblurring process. \citet{chen2019u} introduce a deep-stacked CAE and U-Net structure for deblurring the spatial images impacted by atmospheric turbulence. In a subsequent work of \citet{chen2020blind}, they use a deblurring noise suppression block instead of the convolutional layer in the U-Net structure to eliminate noise and extract more structural features.

\section{Training loss functions}
\label{sec:loss_func}
In image restoration tasks, including image deblurring, training loss functions play a significant role in restoring clearer images with more texture details \cite{zhao2016loss}. Among various choices available, the content loss that measures the difference between the restored outcome and the original target image is the most widely used, and this loss can further improve the quality of the restored image when combined with auxiliary terms \cite{cho2021rethinking}.  Likewise, many studies in the literature combine multiple loss functions in the form of a weighted sum to take advantage of the benefits of various loss functions and enhance the quality of recovered image.  In this section, we list several well-known loss functions and discuss their impacts on the restored outcome. 
For an easier overview, Table~\ref{tab:loss} summarizes training loss functions as well as application types considered in the studies reviewed in this paper.
\begin{table}[t!]
\begin{center} 
\caption{The applications and loss functions of blind deblurring papers in chronological order}
{
\resizebox{\textwidth}{!}{
\begin{tabular}{lccc}
\hline
  & Blur type & Application & Training Loss\\
\hline
\citet{hradivs2015convolutional} & Uniform motion, Defocus & Text &  L2-norm content loss\\
\citet{sun2015learning} & Uniform/Non-uniform motion & General & -\\
\citet{schuler2015learning} & Uniform/Non-uniform motion & General & L2-norm content loss\\
\citet{yan2016blind} & Uniform/Non-uniform motion & General & Cross entropy, L2-norm content loss\\
\citet{chakrabarti2016neural} & Uniform motion& General & L2-norm content loss\\
\citet{gong2017motion} & Uniform/Non-uniform motion & General & Cross entropy\\
\citet{ramakrishnan2017deep} & Non-uniform motion & General & L1-norm content loss, Adversarial loss, Perceptual loss  \\
\citet{nah2017deep} & Dynamic scene (multiple sources) \cite{hyun2013dynamic} & General & L2-norm content loss, Adversarial loss\\
\citet{xu2017motion} & Uniform/Non-uniform motion  & General & Regularized L1-norm content loss\\
\citet{nimisha2017blur} & Uniform/Non-uniform motion & General &  L2/L1-norm content loss, Gradient loss, Adversarial loss\\
\citet{li2018learning} & Uniform/Non-uniform motion & General &  Cross entropy (binary)  \\
\citet{kupyn2018deblurgan} & Non-uniform motion & General &Adversarial loss, Perceptual loss\\
\citet{zhang2018dynamic} & Dynamic Scene (multiple sources) & General & L2-norm content loss \\
\citet{tao2018scale} &  Dynamic Scene (multiple sources) & General & L2-norm content loss\\
\citet{gao2018stacked} & Atmosphere turbulence & Space targets (Remote sensing) & L2-norm content loss\\
\citet{shen2018deep} & Uniform motion & Face & Content loss, Adversarial loss, Perceptual loss\\
\citet{zhang2019deep} & Non-uniform motion & General & L2-norm content loss   \\
\citet{aljadaany2019douglas} & Non-uniform motion & General & L2-norm content loss, Adversarial loss\\
\citet{kupyn2019deblurgan} & Non-uniform motion & General & L2-norm content loss, Perceptual loss, Adversarial loss \\
\citet{gao2019dynamic} & Dynamic scene (multiple sources)& General & L2-norm content loss\\
\citet{li2019deep} & Uniform motion & General & L2-norm content loss\\
\citet{chen2019u} & Atmosphere turbulence & Space targets (Remote sensing) & L2-norm content loss\\
\citet{shen2019human} & Dynamic scene (multiple sources) & General &  L2-norm content loss \\
\citet{cai2020dark} & Dynamic scene (multiple sources) & General & Regularized L1-norm content loss \\
\citet{purohit2020region} & Dynamic scene (multiple sources) & General & -  \\
\citet{lin2020learning} & Uniform motion & Face & L1-norm content loss, Adversarial loss\\
\citet{chen2020blind} & Atmosphere turbulence & Space targets (Remote sensing) & - \\
\citet{zhang2020deblurring} & Dynamic scene (multiple sources) & General& Perceptual loss, L2-norm content loss, Adversarial loss, Relativistic loss\\
\citet{ren2020neural} & Uniform motion & General & L2-norm content loss\\
\citet{asim2020blind} & Uniform/Non-uniform motion & General & L2-norm content loss \\
\citet{xu2021attentive} & Dynamic scene (multiple sources) & General & L2-norm content loss, Gradient loss \\ 
\citet{zhao2021gradient} & Dynamic scene (multiple sources) & General & L1-norm content loss, Gradient loss, Perceptual loss, Adversarial loss\\
\citet{chen2021attention} & Dynamic scene (multiple sources) & General & L2-norm content loss\\
\citet{cho2021rethinking} & Dynamic scene (multiple sources) & General & L1-norm content loss, L1-norm Frequency reconstruction loss\\
\citet{dong2021deep} & Uniform/Non-uniform & General & L2-norm content loss\\
\citet{luo2021bi} & Dynamic scene (multiple sources) & General & L2-norm content loss\\
\citet{li2021single} & Dynamic scene (multiple sources) & General & Ranking content loss, L2-norm content loss, Adversarial Loss\\
\citet{ren2021deblurring} & Dynamic scene (multiple sources) & General & L2-norm content loss \\
\citet{wu2021stack} & Uniform motion & Face & L2-norm content loss\\
\citet{hu2021pyramid} & Dynamic scene (multiple sources) & General & L2-norm content loss\\
\citet{quan2021gaussian} & Defocus & General & L2-norm content loss\\
\citet{tsai2021banet} & Dynamic scene (multiple sources) & General & L2-norm content loss\\
\hline
\label{tab:loss}
\end{tabular} }}
\end{center}
\end{table}

In what follows, we let $I$ denote the ground truth image and $I'$ be either the final restored image ($\hat{I}$) or a generated image in the GAN structure ($G(B)$). We use $K$ to denote the total number of scales when a multi-scale structure is considered. In addition, $N$ represents the total number of pixels. 
\begin{itemize}
    \item \textbf{Content loss (Reconstruction loss)}
    is commonly formulated in two conventional types: L2-norm content loss, or  mean squared error (MSE), and L1-norm content loss, or mean absolute error (MAE). The content loss computes the discrepancy of pixel values between the ground truth image ($I_k$) and an output image of a network according to the corresponding norm \cite{nah2017deep, zhang2019deep, tao2018scale}, and minimizing this loss helps a network restore the overall content and structure of the image.  Some studies prefer using L1-norm since L2-norm tends to lose high frequency information in an image generation process \cite{zhao2021gradient}. In general, the content loss is formulated as
    \begin{equation}
        L_{Cont} = \frac{1}{N}\sum_{k=1}^K \|I'_k - I_k \|_{norm}.
        \label{eq:cont}
    \end{equation}
  
    \item \textbf{Perceptual loss} \cite{johnson2016perceptual}
    compares the ground truth and output images in their CNN feature representations rather than pixel-wise differences as in the content loss. This loss function tries to make an output image perceptually indistinguishable from the ground truth image while the content loss sometimes produces over-smooth pixels and blurry artifacts \cite{kupyn2018deblurgan}. As such, the perceptual loss is a good alternative that can overcome some drawbacks of the content loss.  The perceptual loss is defined as
    \begin{equation}
        L_{Perc} = \frac{1}{C_j H_j W_j}\| \phi_j(I') - \phi_j(I)\|_2
    \end{equation}
    where $\phi_j$ is generally the feature map resulting from the activation of the $j$th convolutional layer of VGG19 network \cite{simonyan2014very}, a pre-trained network for generating feature maps.  The activation of the $j$th layer produces a feature map of size $W_j \times H_j \times C_j$ where $C_j$, $W_j$, and $H_j$ denote the number of channels, width, and height of the corresponding feature map, respectively.  The \textit{cov3\_3} feature maps of VGG19 is commonly selected to compute the loss along with the Euclidean distance (L2-norm).  The feature maps of later layers, such as \emph{cov3\_3}, tend to have more prominent information than earlier layers producing readily recognizable features \cite{zhao2021gradient}.
    \item \textbf{Regularized content loss} includes a regularization term in addition to the general content loss to enforce sparsity on image priors for better restoration outcomes \cite{cai2020dark}. The regularized content loss function is formulated as
    \begin{equation}
        L_{RC} = \sum_{k=1}^K \|I'_k - I_k \|_{norm} + \lambda P(I'_k)
        \label{eq:Reg}
    \end{equation}
    where $P(\cdot)$ denotes some prior information of the generated image \cite{xu2017motion}. The prior could be image gradients \cite{xu2017motion} or more practical information, such as dark channel or bright channel \cite{cai2020dark}.   
    
    \item \textbf{Adversarial loss} \cite{goodfellow2014generative} is employed to generate realistic images in a GAN structure \cite{nah2017deep}. Although it is commonly used for the generator network in a GAN structure, other deep structures can also adopt this loss to improve training procedures.  For instance, the discriminator architecture in \citet{radford2015unsupervised} is also trained by using this loss to classify if the generated latent image ($I'$) is a blurred or sharp image \cite{nah2017deep}. The adversarial loss seeks to restore more texture details of the output while content and perceptual losses focus on the "macro-structure" of the restored image \cite{kupyn2018deblurgan}.  The adversarial loss is computed as 
    \begin{equation}
        L_{Adv} = E_{I \sim p_{target}(I)}[log(D(I))] + E_{B \sim p_{blurred}(B)}[log(1 - D(G(B)))],
        \label{eq:adv}
    \end{equation}
    where $p_{target}$ and $p_{blurred}$ are the distributions of the ground truth image and the blurred image, respectively. 
    \item \textbf{Gradient loss} can effectively preserve the edges in the output images and recover sharper images \cite{nimisha2017blur}. Image gradients typically contain significant information about the texture details and edges, but severe blur can make the edges indistinguishable.  The gradient loss can help retrieve more salient edges during a training process by imposing sparser gradient difference \cite{xu2021attentive, zhao2021gradient}. The gradient loss can be separated into terms associated with vertical and horizontal gradients as directional gradient loss, as shown in
    \begin{equation}
        L_{Grad} = \|(\nabla I' - \nabla I)_x \|_{norm} + \|(\nabla I' - \nabla I)_y \|_{norm}
        \label{eq:gradient}
    \end{equation}
    
    where $\nabla$ is the gradient operator.
     
    \item \textbf{Frequency content loss} measures the discrepancy between the output and target image in the frequency domain.  In general, a deblurring process tries to retrieve high-frequency components that are lost as a consequence of blur \cite{cho2021rethinking}, so using this loss can help restore an image with more clarity. For the computation, both output and target images are mapped into the frequency domain by using fast Fourier transform (FFT), denoted by $F(\cdot)$, and the frequency content loss is calculated as (summed over multiple scales)
    \begin{equation}
        L_{FR} = \sum_{k=1}^K \|F(I'_k) - F(I_k) \|_{norm}.
        \label{eq:frequency}
    \end{equation}
    
    \item \textbf{Ranking content loss} \cite{zhang2019ranksrgan} is originally proposed for an image super-resolution process to train the generator network in a GAN structure. The loss is computed by a trained Siamese network \cite{zagoruyko2015learning, bromley1993signature} which is designed to evaluate image quality. As shown in Fig.~\ref{fig:Siamese}, Siamese network itself consists of two parallel branches with the exact same structure and shared weights \cite{zagoruyko2015learning}.  This network takes two same images with different quality ($x_1$ and $x_2$) as inputs and is trained based on margin-ranking loss, defined as
    \begin{equation}
        L_{MR} = \max \left(0, \gamma (R(x_1) - R(x_2)) + \epsilon \right),
        \label{eq:margin}
    \end{equation}
    which is widely used in sorting problems \cite{zhang2019ranksrgan}.  $\gamma$ represents the quality criterion having 1 if $R(x_1)$ is greater than $R(x_2)$; otherwise, -1. $R(\cdot)$ is the Siamese network whose output specifies the ranking scores of image pairs \cite{zhang2019ranksrgan}. $\epsilon$ is determined arbitrarily to control the quality scores between the outputs from the two branches. Once the Siamese network is trained, it takes a recovered image as an input and computes the corresponding ranking score according to 
    \begin{equation}
        L_{RC} = R(I').
        \label{eq:rank}
    \end{equation}
    Hence, this loss seeks to reduce the discrepancy between the deblurred image score and target image score during the training process based on the trained Siamese network. 
    \begin{figure}[!h]
    \centering
    \captionsetup{justification=centering}
     \includegraphics[width=15cm]{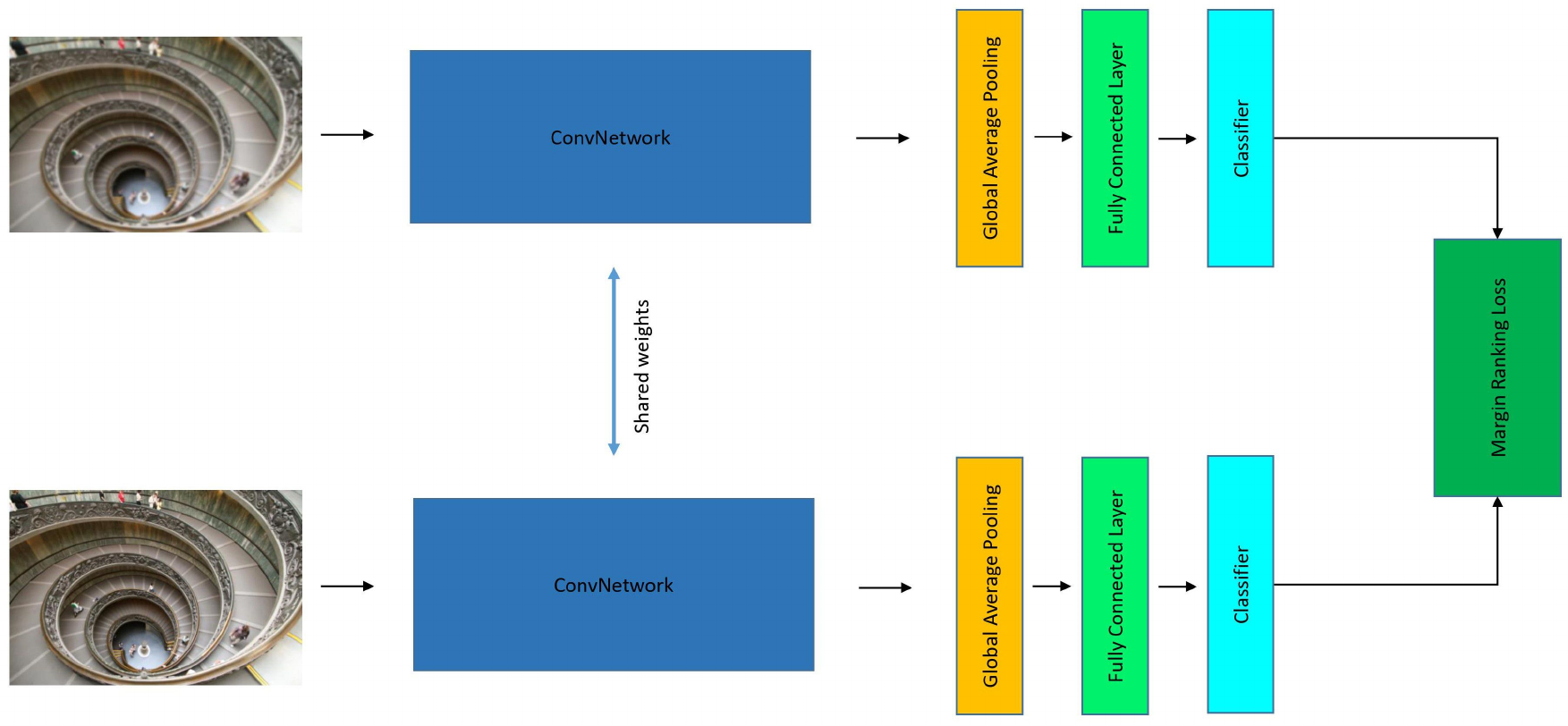}
    \caption{Siamese network, a ConvNetwork consisting of several convolutional blocks, each of which has convolutional layer, batch normalization, and LeakyRelu function.  }
    \label{fig:Siamese}
    \end{figure}
     
    \item \textbf{Relativistic loss} \cite{zhang2020deblurring} is developed for the relativistic GAN \cite{jolicoeur2018relativistic} in which the discriminator estimates the probability that the real data is more realistic than the fake data (randomly sampled), by reformulating the adversarial loss as
     \begin{equation}
        L_{RL} = - \left[\log(\sigma(C(I) - E(C(I')))) + \log(1 - (\sigma(I') - E(C(I)))) \right]
        \label{eq:Rloss}
    \end{equation}
   
    where $\sigma(\cdot)$ and $C(\cdot)$, respectively, are the sigmoid function and the prior-activated feature representation of discriminator network, respectively. $E(\cdot)$ denotes the averaging operation of images in a single batch. This loss function can help restore a more realistic image for the output of GAN structure \cite{zhang2020deblurring}.
    \item \textbf{Cross entropy} is used for deblurring classification networks that predict the probability of the input image being blurred.  The probability is computed by applying a sigmoid activation function to the last layer of a network under consideration \cite{li2018learning}.  The cross entropy loss function is defined as
    \begin{equation}
        L_{CE} = - \sum_{i=1}^{N_T}  y'_i  \text{log}(\hat{y_i})
        \label{eq:cross}
    \end{equation}
    where $N_T$ is the total number of images in training data, $y'_i$ and $\hat{y_i}$ denote the target label and the probability output of the network, respectively.  This loss function can also be used for a deblurring process where a network predicts the probability of movements in the horizontal and vertical directions to estimate motion flow as done in \cite{gong2017motion}.  
\end{itemize}

\section{Image Deblurring Datasets}
\label{sec:deblur_dataset}
This section describes widely-used datasets in the literature of both prior-based optimization and deep neural image deblurring methods. We also discuss some domain-specific datasets that are used in some application-based works. 
\begin{figure}[!b]
    \centering
    \begin{subfigure}[h]{.23\textwidth}
        \centering
        \includegraphics[width=\textwidth]{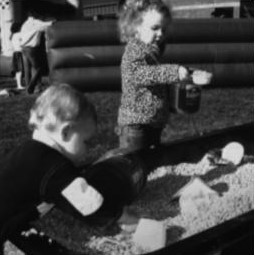}
    \end{subfigure} \vspace{2mm}%
    ~
    \begin{subfigure}[h]{.23\textwidth}
        \centering
        \includegraphics[width=\textwidth]{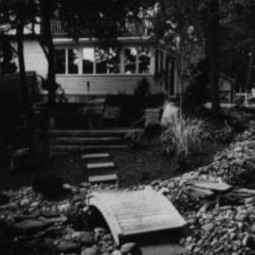}
    \end{subfigure}
    ~
     \begin{subfigure}[h]{.23\textwidth}
        \centering
        \includegraphics[width=\textwidth]{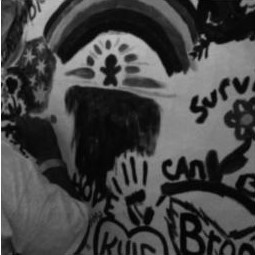}
    \end{subfigure}
    ~
    \begin{subfigure}[h]{.23\textwidth}
        \centering
        \includegraphics[width=\textwidth]{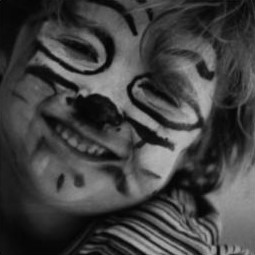}
    \end{subfigure}
    ~
    \begin{subfigure}[h]{.23\textwidth}
        \centering
        \includegraphics[width=\textwidth]{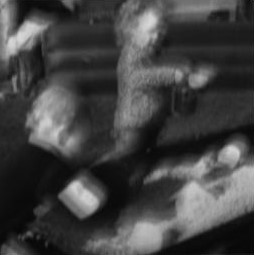}
    \end{subfigure} \vspace{2mm}%
    ~
    \begin{subfigure}[h]{.23\textwidth}
        \centering
        \includegraphics[width=\textwidth]{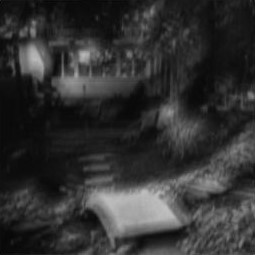}
    \end{subfigure}
    ~
     \begin{subfigure}[h]{.23\textwidth}
        \centering
        \includegraphics[width=\textwidth]{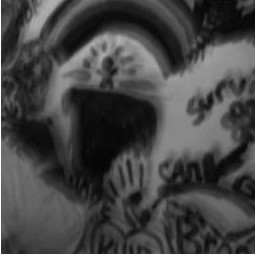}
    \end{subfigure}
    ~
    \begin{subfigure}[h]{.23\textwidth}
        \centering
        \includegraphics[width=\textwidth]{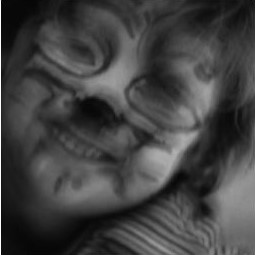}
    \end{subfigure}
    \caption{\citet{levin2009understanding} dataset instances where the first row shows original images and the second row shows blurred images.}
    \label{fig:levin_dataset}
\end{figure}

\begin{figure}[t!]
    \centering
    \begin{subfigure}[h]{.23\textwidth}
        \centering
        \includegraphics[width=\textwidth]{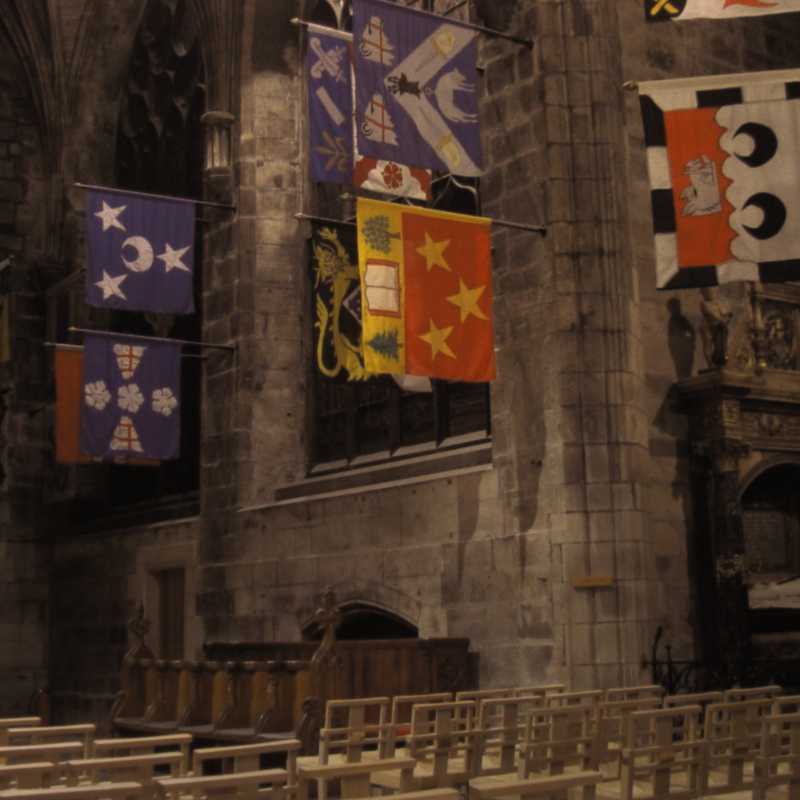}
    \end{subfigure} \vspace{2mm}%
    ~
    \begin{subfigure}[h]{.23\textwidth}
        \centering
        \includegraphics[width=\textwidth]{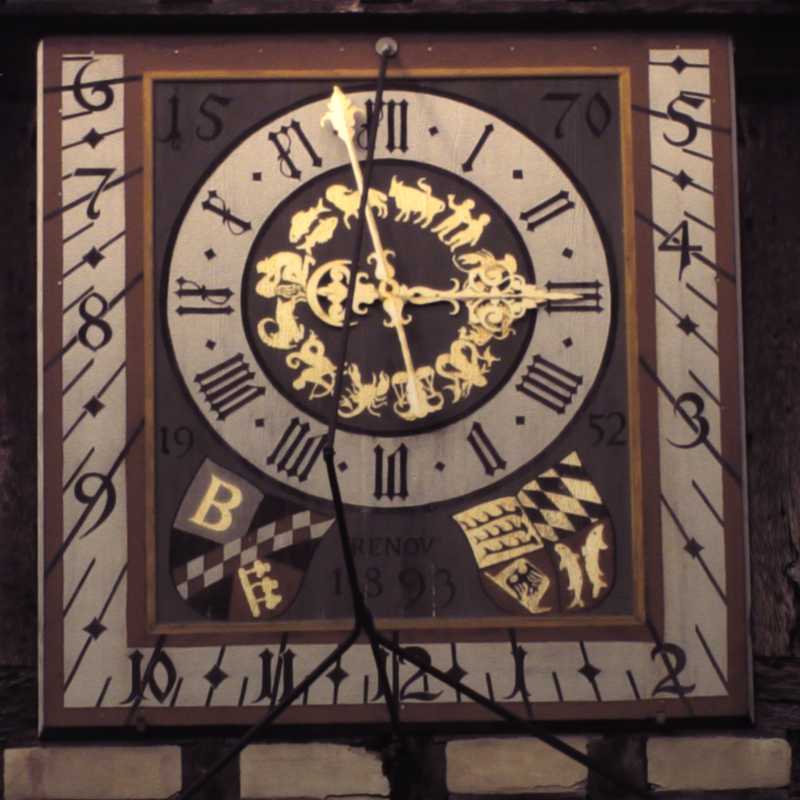}
    \end{subfigure}
    ~
     \begin{subfigure}[h]{.23\textwidth}
        \centering
        \includegraphics[width=\textwidth]{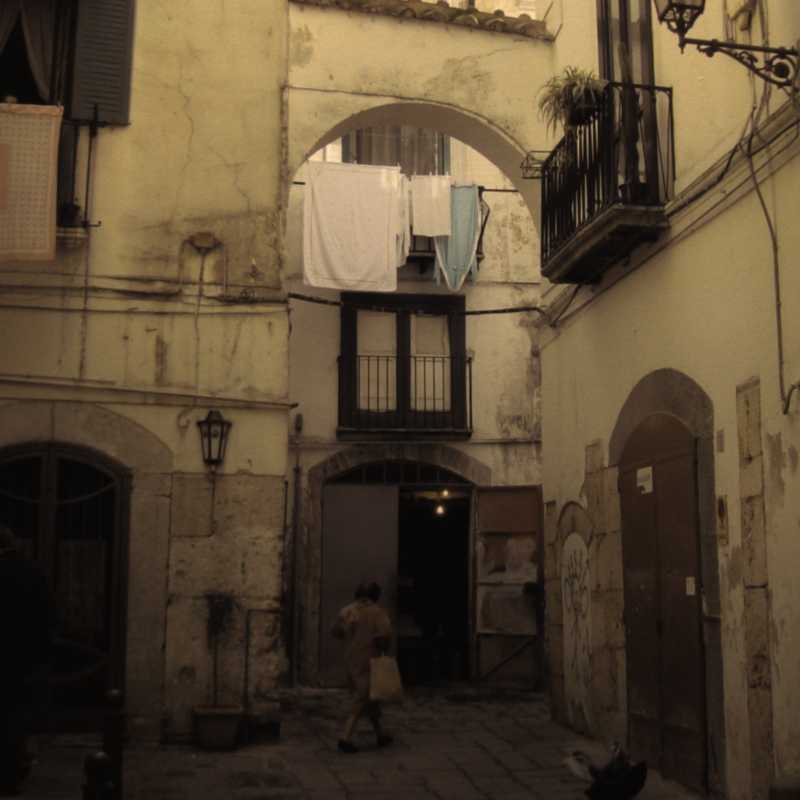}
    \end{subfigure}
    ~
    \begin{subfigure}[h]{.23\textwidth}
        \centering
        \includegraphics[width=\textwidth]{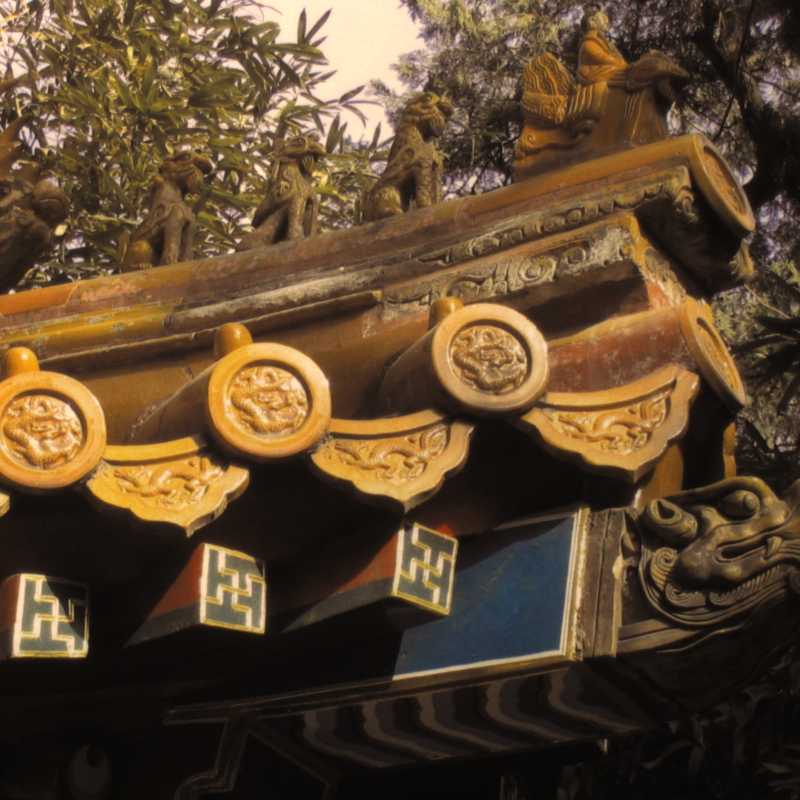}
    \end{subfigure}
    ~
    \begin{subfigure}[h]{.23\textwidth}
        \centering
        \includegraphics[width=\textwidth]{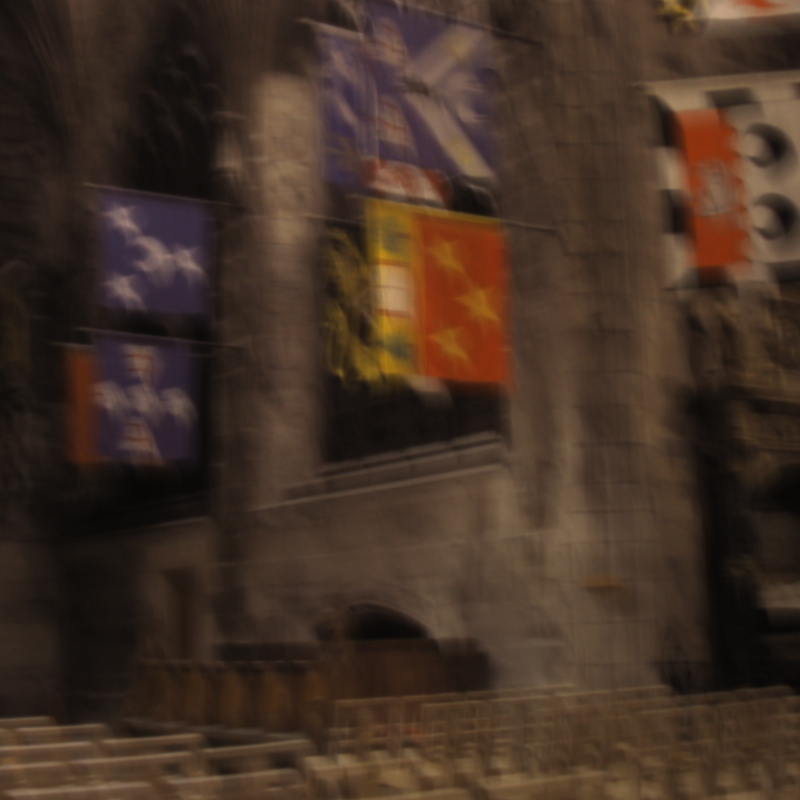}
    \end{subfigure} \vspace{2mm}%
    ~
    \begin{subfigure}[h]{.23\textwidth}
        \centering
        \includegraphics[width=\textwidth]{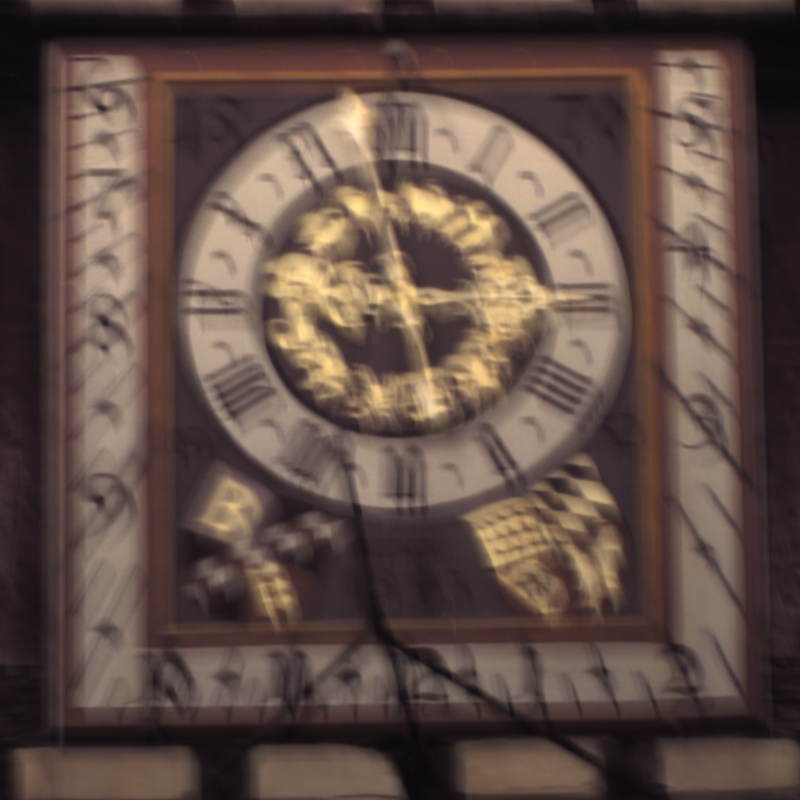}
    \end{subfigure}
    ~
     \begin{subfigure}[h]{.23\textwidth}
        \centering
        \includegraphics[width=\textwidth]{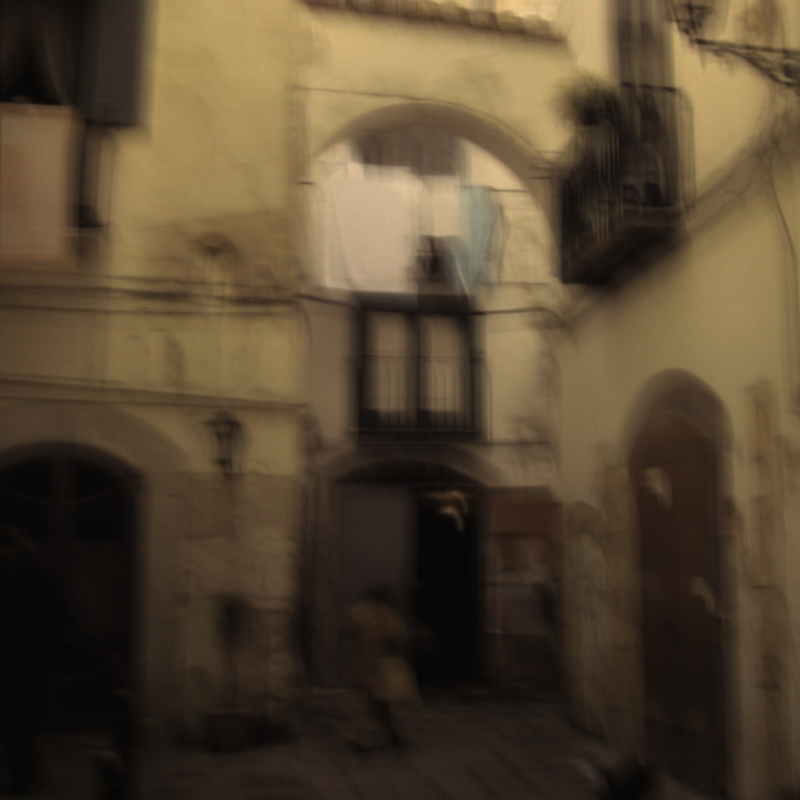}
    \end{subfigure}
    ~
    \begin{subfigure}[h]{.23\textwidth}
        \centering
        \includegraphics[width=\textwidth]{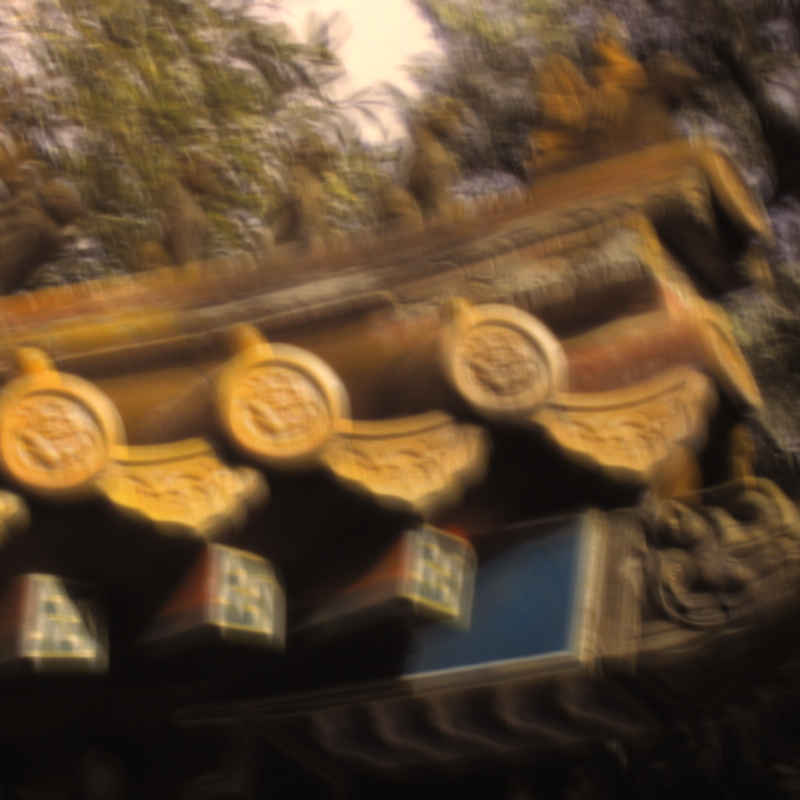}
    \end{subfigure}
    \caption{\citet{kohler2012recording} dataset instances where the first row shows original images and the second row shows blurred images.}
    \label{fig:kohler_dataset}
\end{figure}

\begin{itemize}
\item \textbf{\citet{levin2009understanding} dataset} includes solely uniform blurred images. It has a total of 32 blurred images generated by 4 gray-scale images with 8 uniform kernels.  Since this dataset does not involve any case with non-uniform blur kernels (the type of blurriness found in real-world situations), its usage is quite limited.  In addition, it has a lack of diversity in the type of scenes and the size of images \cite{lai2016comparative}. Figure \ref{fig:levin_dataset} illustrates the real/blurred images of \citet{levin2009understanding} dataset.

\item \textbf{\citet{kohler2012recording} dataset} has 48 blurred images in total that are generated by applying 12 distinct non-uniform blur kernels to 4 original images. These non-uniform blur kernels are created referring to the results of recording the 6D trajectories of camera motion and simulating the real effects in a lab environment. Also, the effect of real camera shake was examined by simulating the actual procedures of photo shooting with long exposure time. The  instances of the \citet{kohler2012recording} dataset are shown in Figure \ref{fig:kohler_dataset}.

\item \textbf{\citet{sun2013edge} dataset} applies the same eight uniform blur kernels introduced in \citet{levin2009understanding} to a broader set of real scenes (80 images) resulting in a total of 640 blurred images. 

\item \textbf{\citet{lai2016comparative} dataset} applies spatially varying blur kernels as well as uniform (stationary) blur kernels to 25 real-world images. To generate the spatially varying blur kernels, they acquire the real 6D trajectories of camera from cellphone sensors.  For the uniform blur kernels, they use a random selection of the 6D camera trajectories.  They apply 8 kernels (uniform and non-uniform) to 25 latent images, add 1\% Gaussian noise to simulate the camera noise, and create overall 200 uniform/non-uniform synthetic blurred images. Furthermore, \citet{lai2016comparative} consider 100 more real blurred images which were taken under various settings and circumstances. Figure \ref{fig:lai_dataset} displays several blurred images of this dataset.  
\begin{figure}[!b]
    \centering
    \begin{subfigure}[h]{3.75 cm}
        \centering
        \includegraphics[width=3.75 cm]{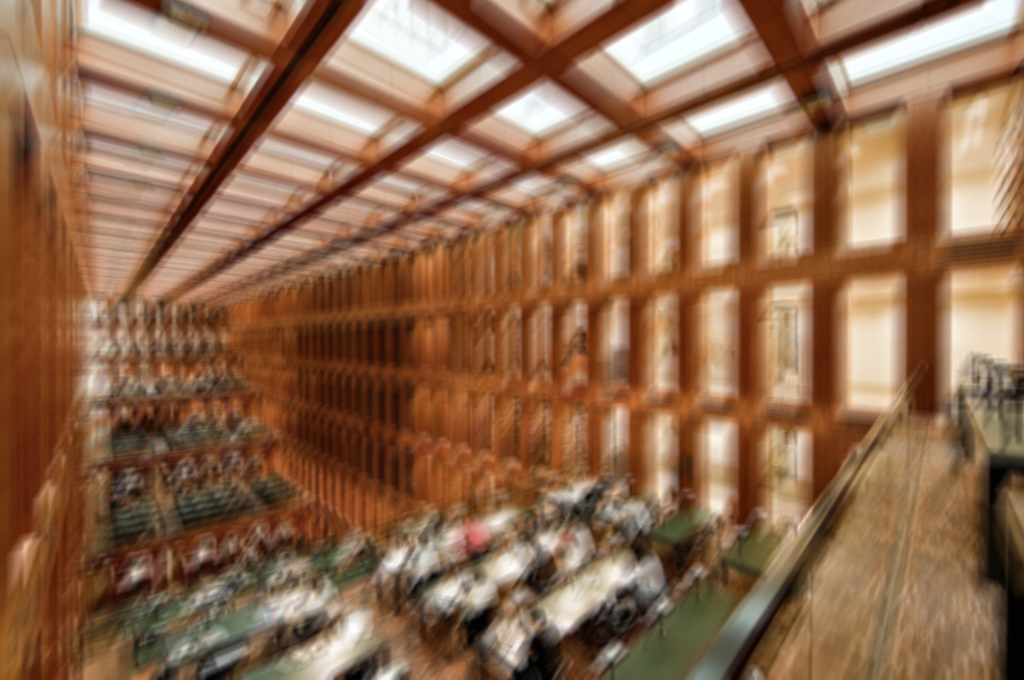}
    \end{subfigure} \vspace{2mm}%
    ~
    \begin{subfigure}[h]{3.75 cm}
        \centering
        \includegraphics[width=3.75 cm]{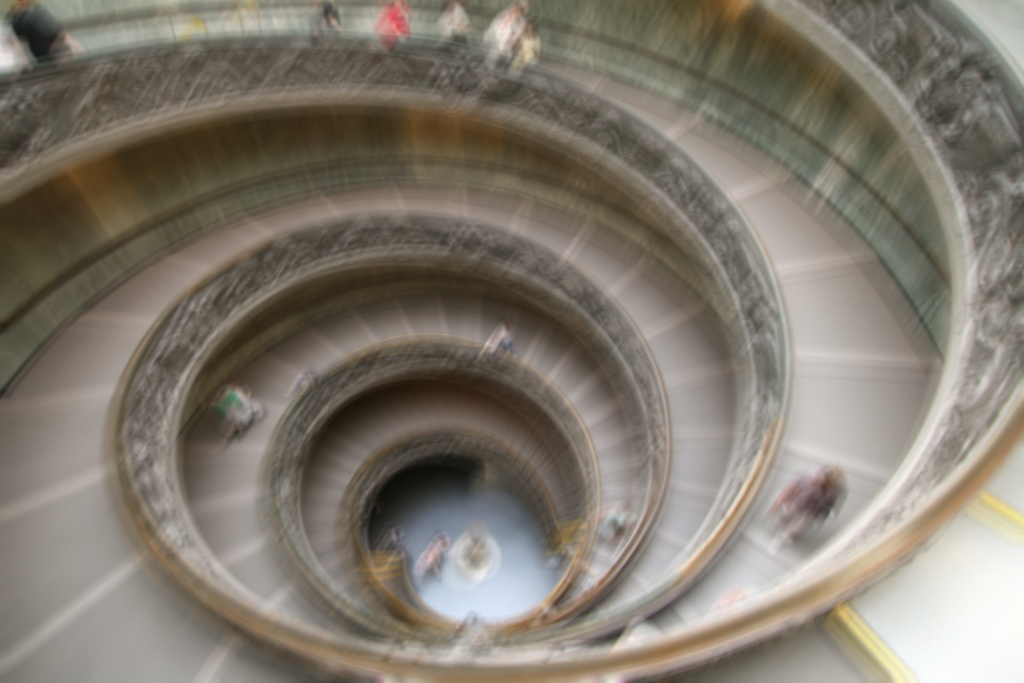}
    \end{subfigure}
    ~
     \begin{subfigure}[h]{3.75 cm}
        \centering
        \includegraphics[width=3.75 cm]{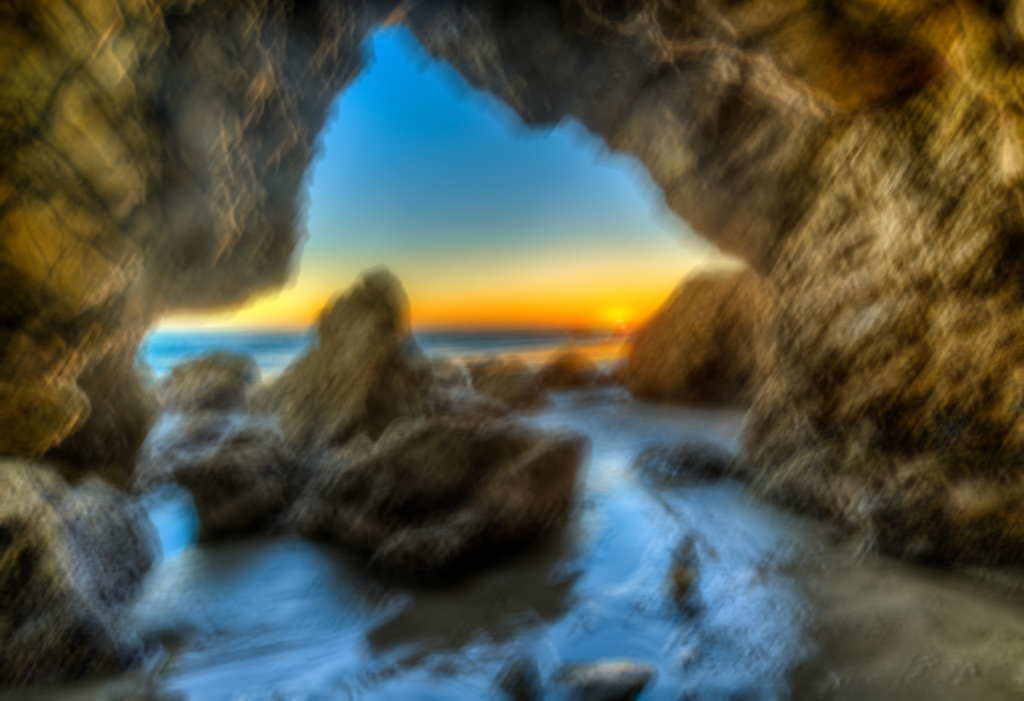}
    \end{subfigure}
    ~
    \begin{subfigure}[h]{3.75 cm}
        \centering
        \includegraphics[width=3.75 cm]{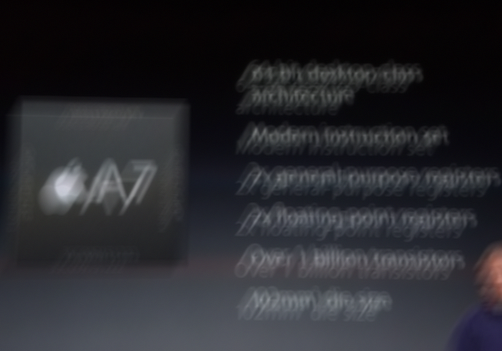}
    \end{subfigure}
    ~
    \begin{subfigure}[h]{3.75 cm}
        \centering
        \includegraphics[width=3.75 cm]{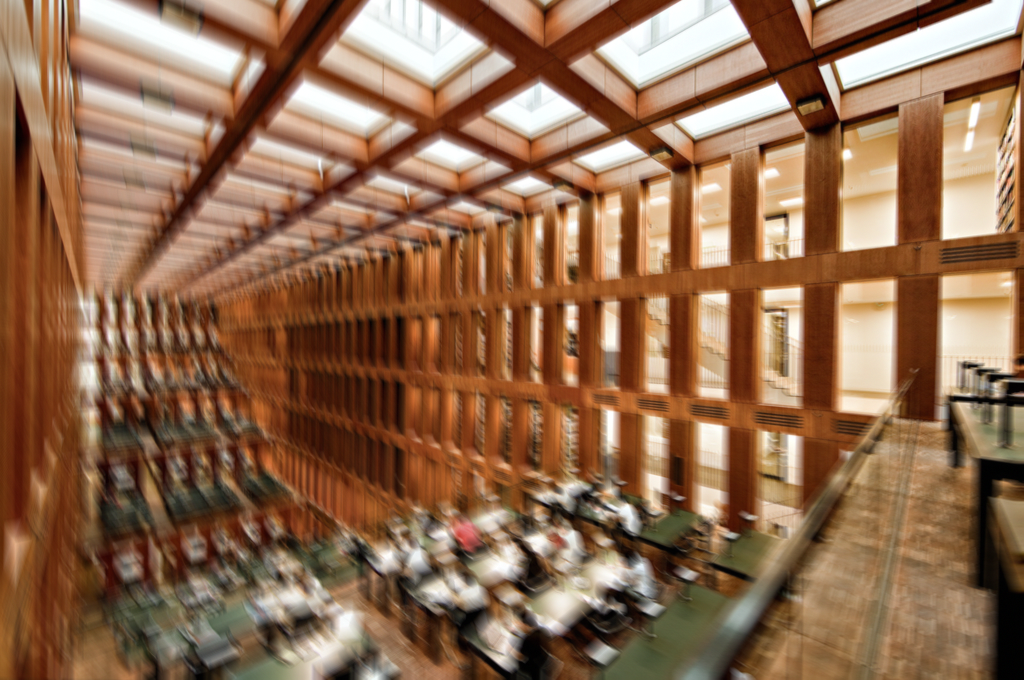}
    \end{subfigure} \vspace{2mm}%
    ~
    \begin{subfigure}[h]{3.75 cm}
        \centering
        \includegraphics[width=3.75 cm]{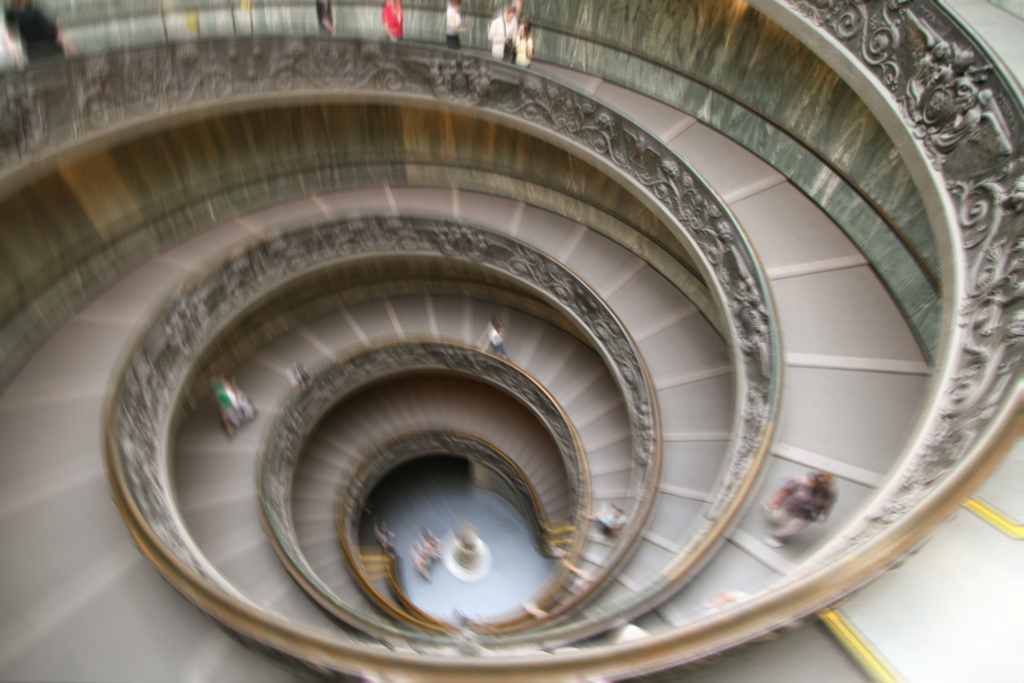}
    \end{subfigure}
    ~
     \begin{subfigure}[h]{3.75 cm}
        \centering
        \includegraphics[width=3.75 cm]{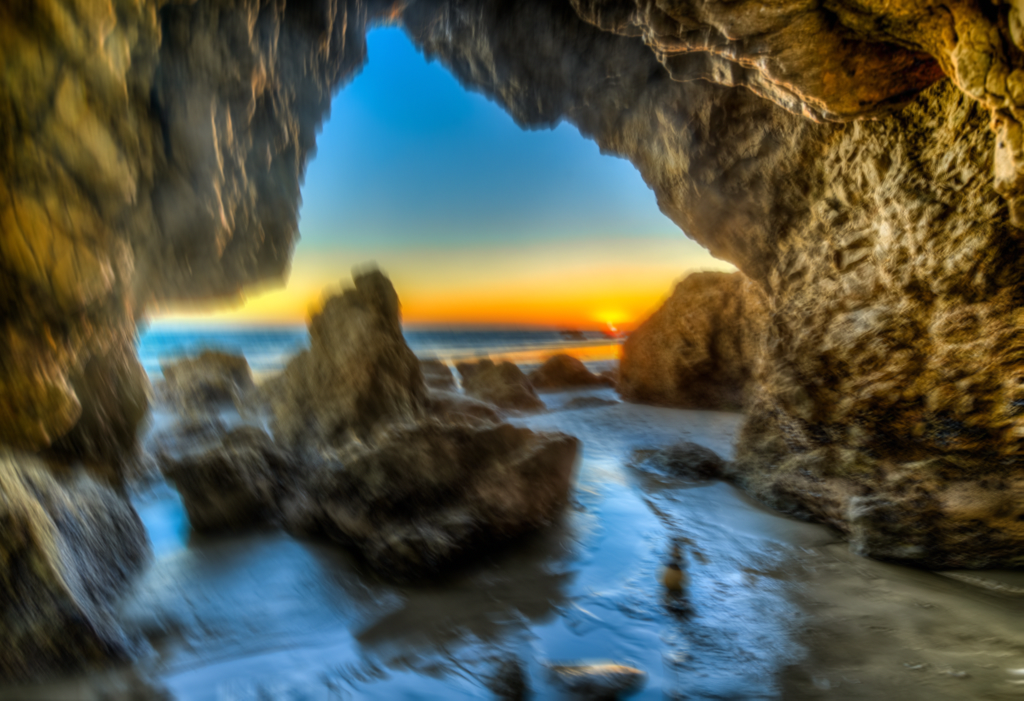}
    \end{subfigure}
    ~
    \begin{subfigure}[h]{3.75 cm}
        \centering
        \includegraphics[width=3.75 cm]{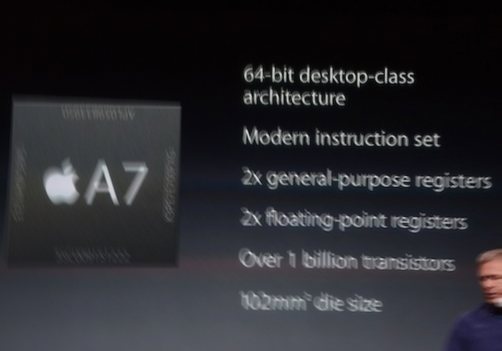}
    \end{subfigure}
    ~
    \begin{subfigure}[h]{3.75 cm}
        \centering
        \includegraphics[width=3.75 cm]{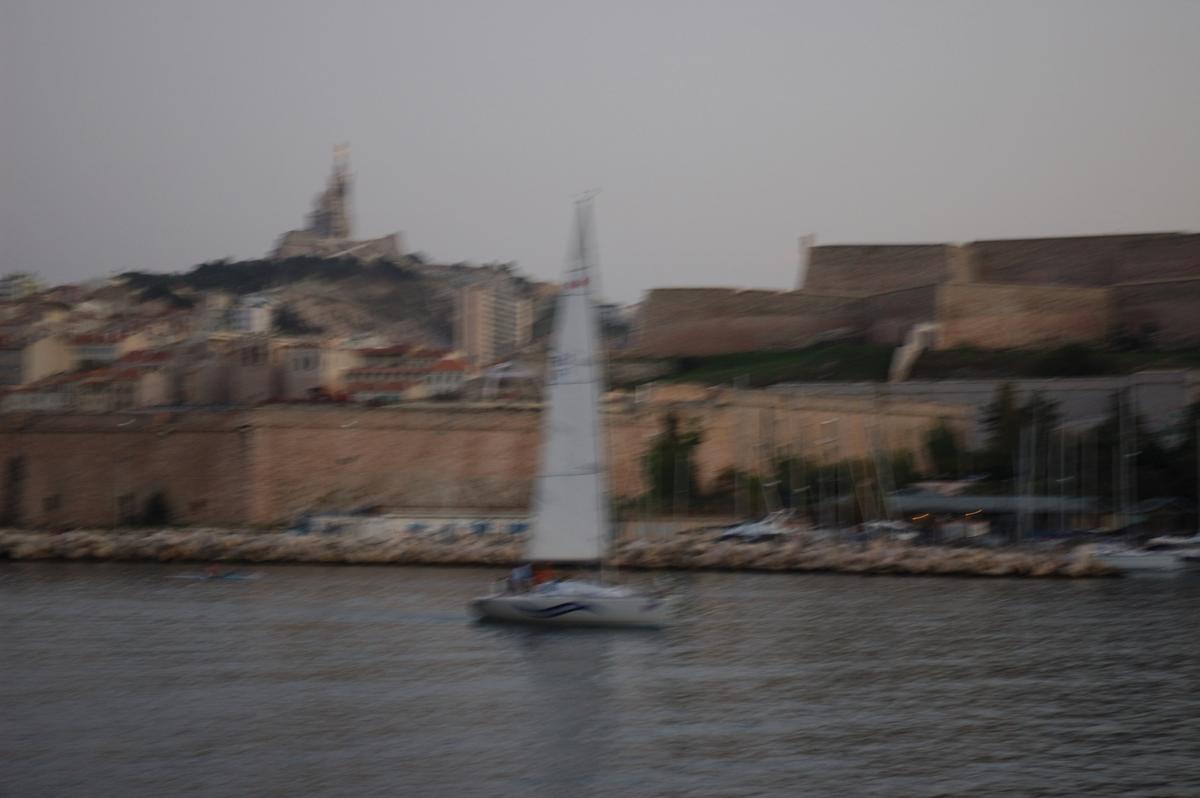}
    \end{subfigure} \vspace{2mm}%
    ~
    \begin{subfigure}[h]{3.75 cm}
        \centering
        \includegraphics[width=3.75 cm]{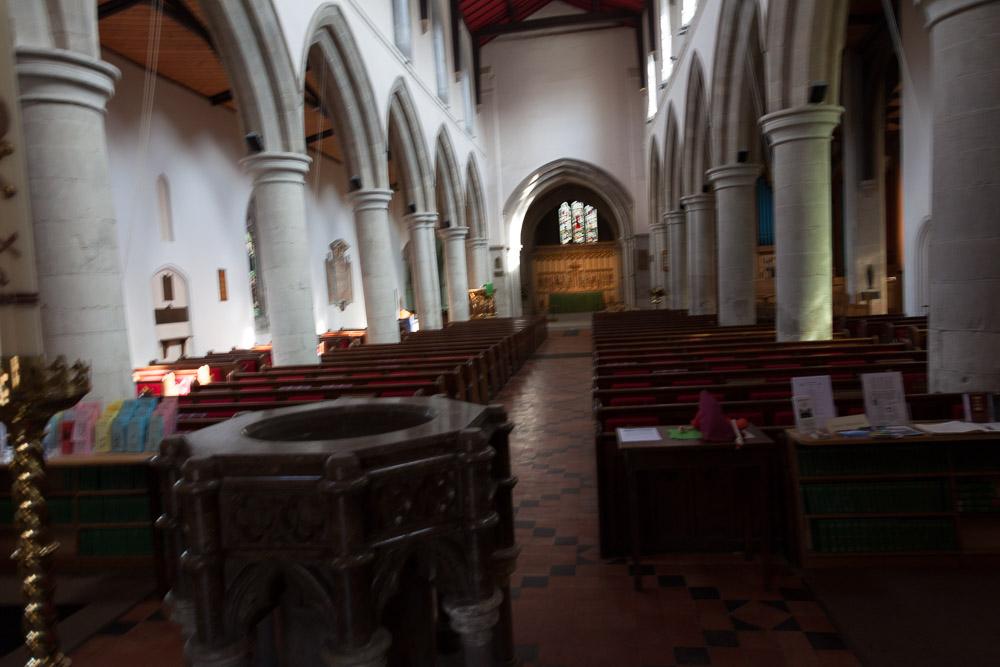}
    \end{subfigure}
    ~
     \begin{subfigure}[h]{3.75 cm}
        \centering
        \includegraphics[width=3.75 cm]{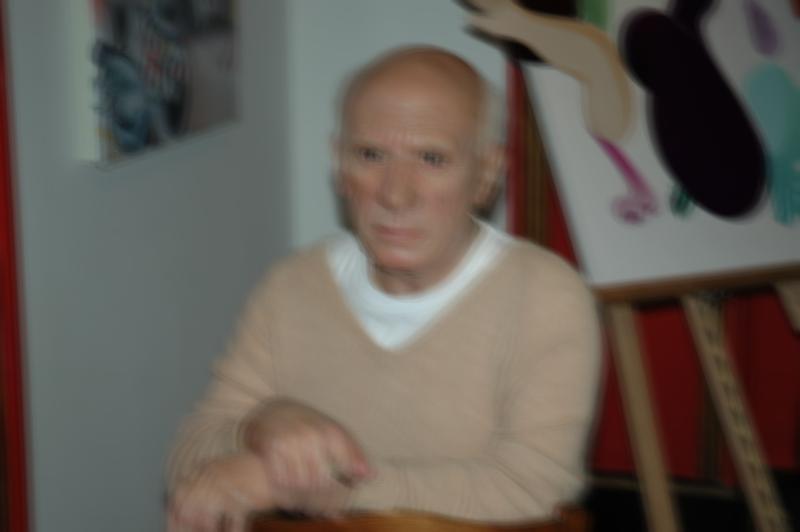}
    \end{subfigure}
    ~
    \begin{subfigure}[h]{3.75 cm}
        \centering
        \includegraphics[width=3.75 cm]{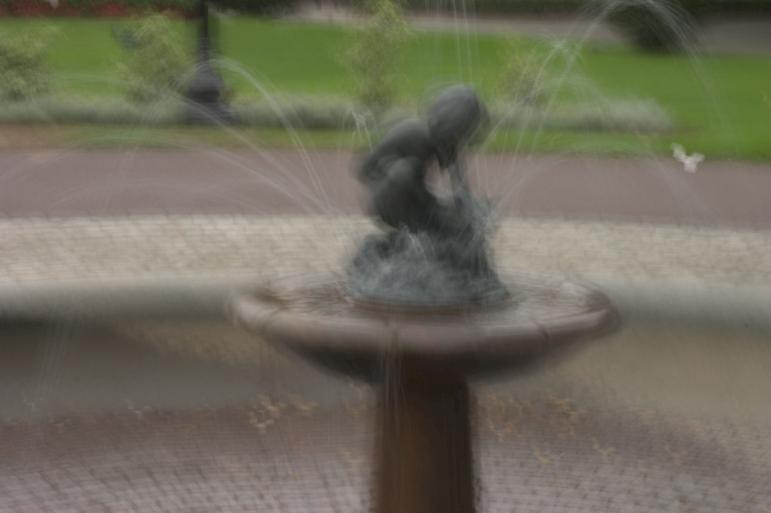}
    \end{subfigure}
    \caption{\citet{lai2016comparative} dataset instances where the first row shows some synthetic uniform blurred images, the second row includes synthetic non-uniform blurred images, and the third row displays real blurred images.}
    \label{fig:lai_dataset}
\end{figure}

\item \textbf{DeepVideoDeblurring (DVD) dataset \cite{su2017deep}} consists of 6,708 blurred frames taken out of 71 videos with their corresponding sharp images. To generate this dataset, real clear videos are recorded by various devices and blurred by applying a longer exposure that is approximately generated by aggregating several short exposures. Furthermore, the dataset is expanded by flipping, rescaling, and rotating the existing blurred frames, ultimately creating 2,146,560 pairs of random cropped patches.  In general, 61 videos with their corresponding patches are used for training, and the remaining videos are used for testing \cite{hu2021pyramid, su2017deep}.   

\item \textbf{GoPro dataset} \cite{nah2017deep} consists of 3,214 pairs of blurry/clear images with the resolution of 1280$\times$720 which are commonly split into 2,103 images for training and 1,111 images for testing \cite{hu2021pyramid, nah2017deep}. They take videos with GOPRO camera and average multiple successive frames \cite{hirsch2011fast} to generate various blurred images. Hence, a mid-frame image is regarded as the ground-truth image of the corresponding synthetic blurred image.  Several blurred images of the GoPro dataset are shown in Figure \ref{fig:GoPro_dataset}. 

\item \textbf{HIDE \cite{shen2019human} dataset} includes complicated blurred images and is generated from diverse scenes, including wide-range and close-range scenes with significant foreground moving objects which the GoPro dataset \cite{nah2017deep} is lack of. To generate blurred images, they average 11 sequential frames of video and take the middle frame as the target image for the corresponding blurry image.  The dataset includes 4,202 scattered people and 4,220 crowded people in terms of the population of the images, and 1,304 long-shot and 7,118 close-ups in terms of object depth.        

\item \textbf{\citet{rim2020real} (RealBlur) dataset} includes a total of 4,738 pairs of blurred/ground-truth images which are generated by an image acquisition system with further post-processing, such as geometric alignment and photometric alignment, to produce realistic blurred images. Their experiments demonstrate this realistic blurred dataset, when used for training, can improve the performance of deep neural structures for real-world blurred images.      
\end{itemize}

\begin{figure}[!t]
    \centering
    \begin{subfigure}[h]{.32\textwidth}
        \centering
        \includegraphics[width=\textwidth]{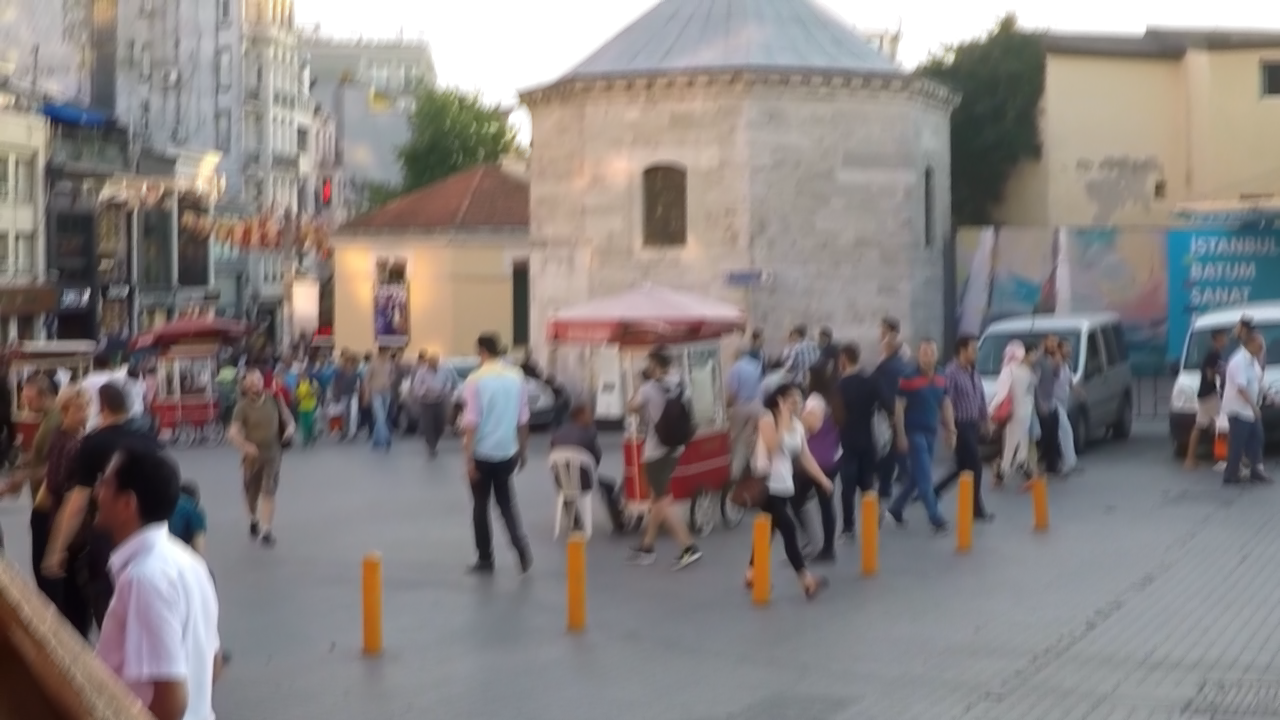}
    \end{subfigure} \vspace{2mm}%
    ~
    \begin{subfigure}[h]{.32\textwidth}
        \centering
        \includegraphics[width=\textwidth]{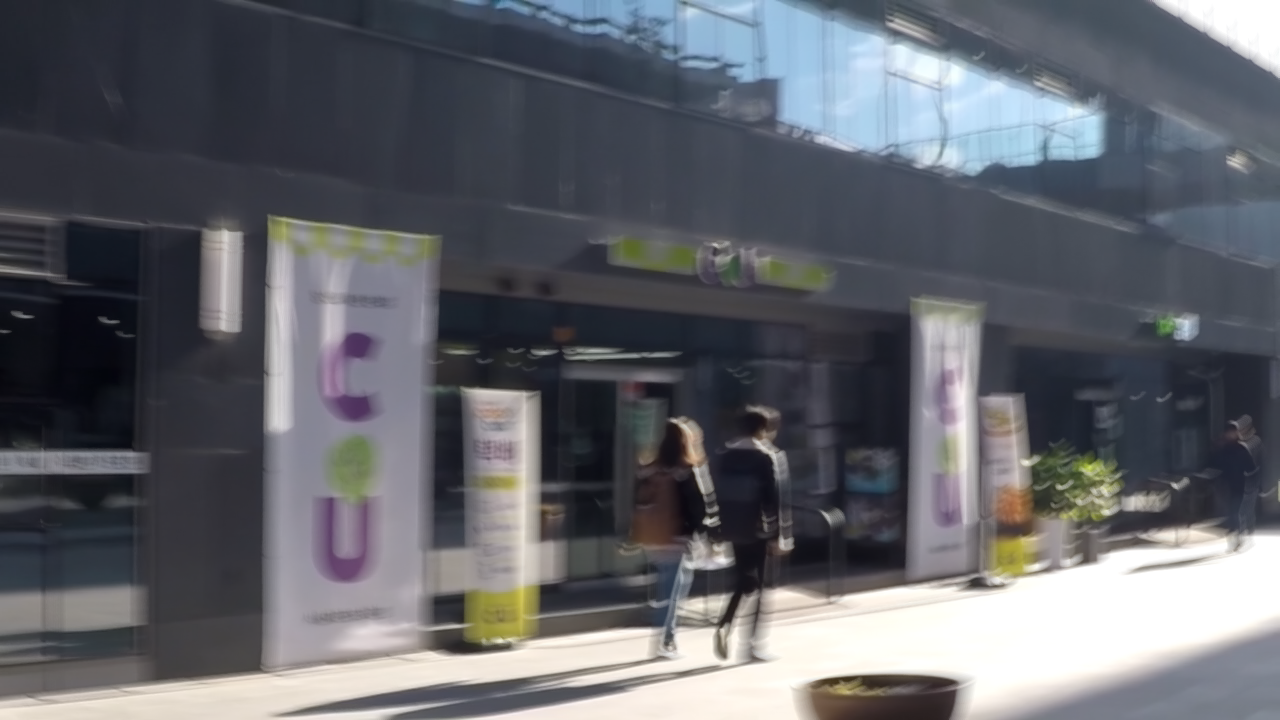}
    \end{subfigure}
    ~
     \begin{subfigure}[h]{.32\textwidth}
        \centering
        \includegraphics[width=\textwidth]{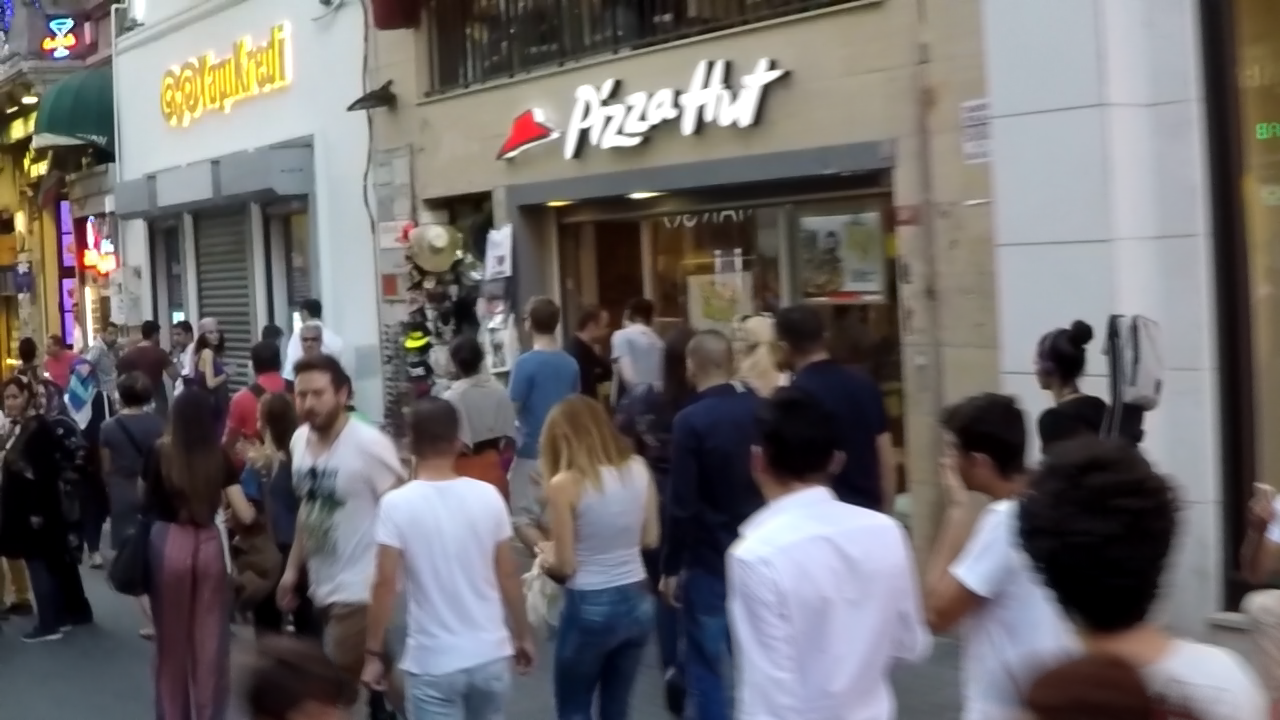}
    \end{subfigure}
    ~
    \begin{subfigure}[h]{.32\textwidth}
        \centering
        \includegraphics[width=\textwidth]{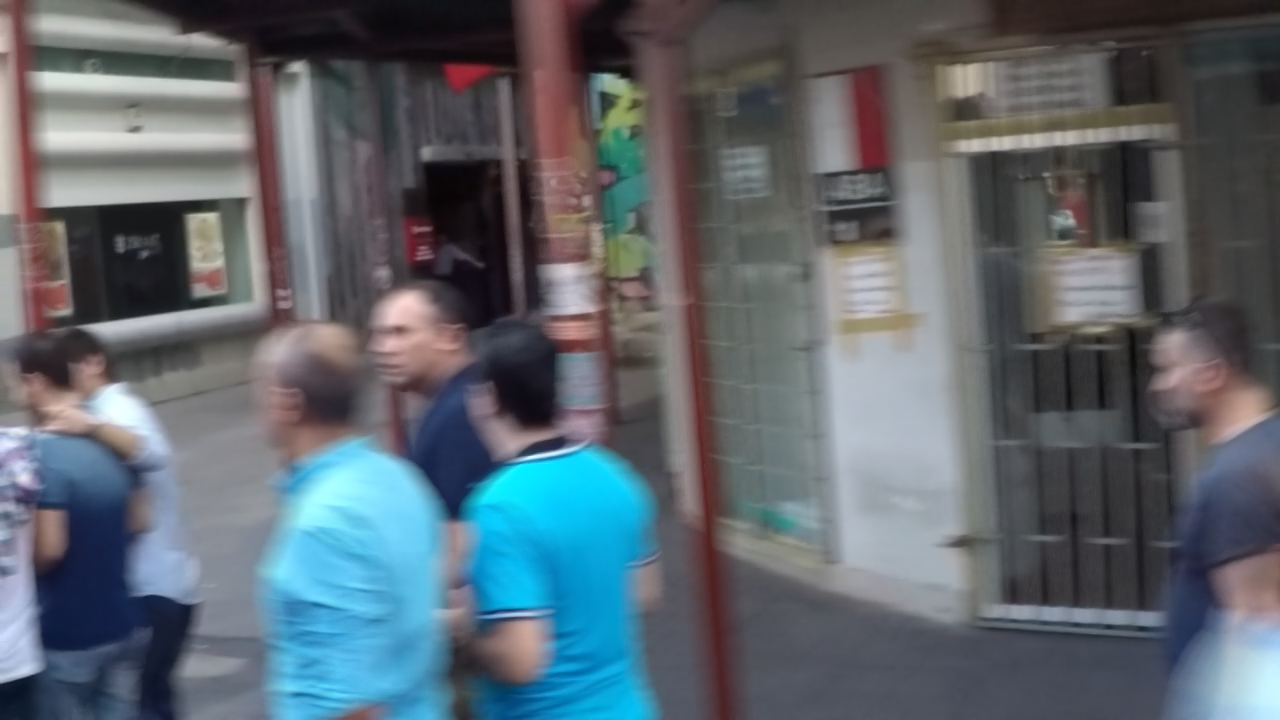}
    \end{subfigure}
    ~
    \begin{subfigure}[h]{.32\textwidth}
        \centering
        \includegraphics[width=\textwidth]{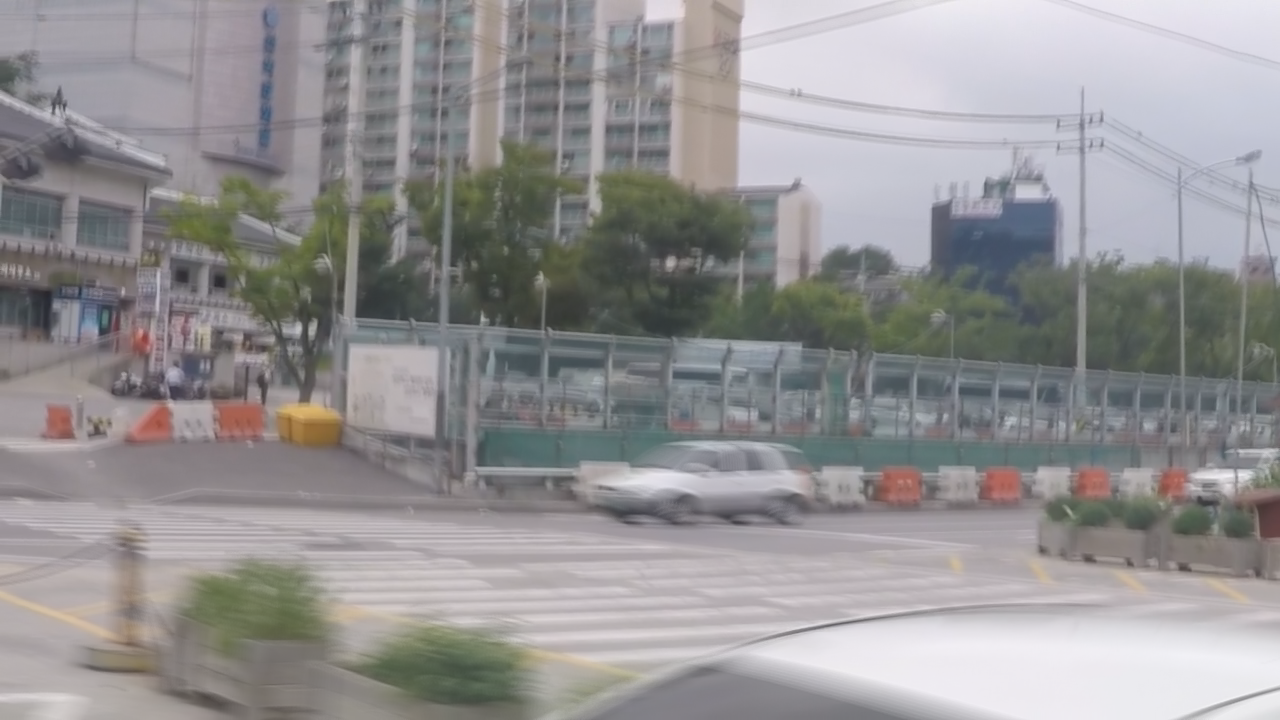}
    \end{subfigure} \vspace{2mm}%
    ~
    \begin{subfigure}[h]{.32\textwidth}
        \centering
        \includegraphics[width=\textwidth]{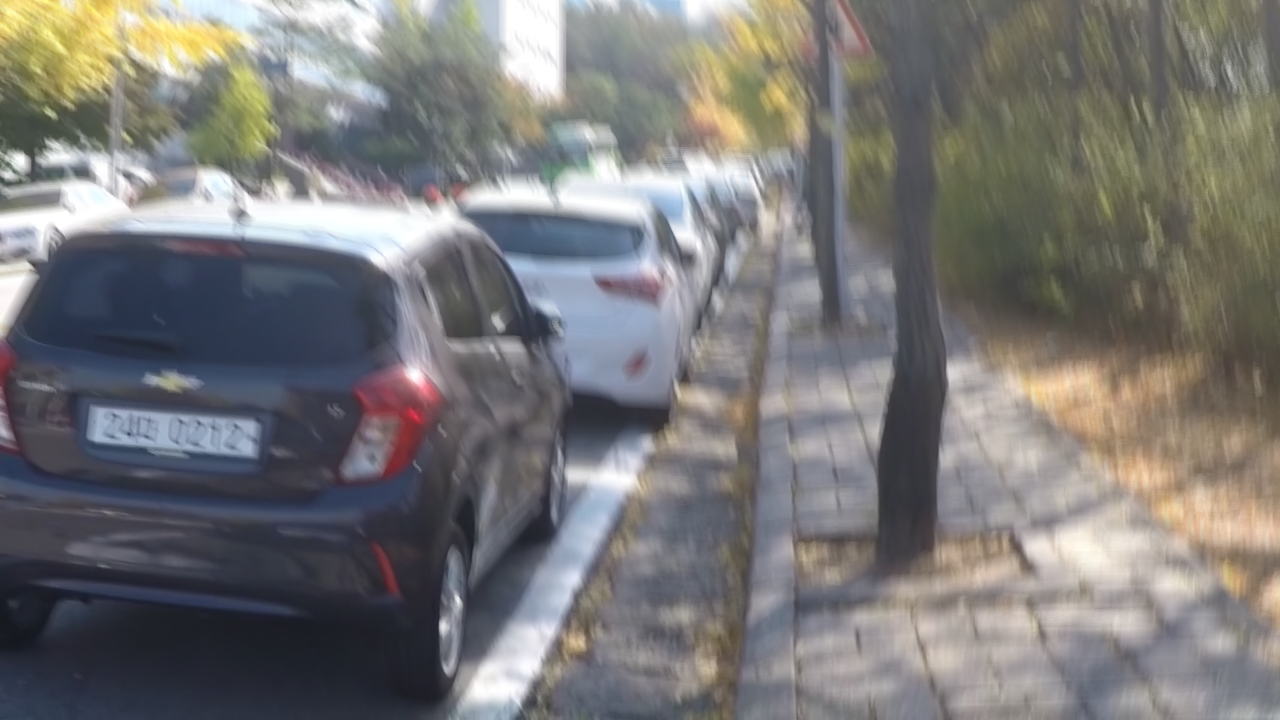}
    \end{subfigure}
    \caption{  GoPro dataset \cite{nah2017deep} instances }
    \label{fig:GoPro_dataset}
\end{figure}

\par
Although most studies use general real-world scenes and images to evaluate their methods, there are also application specific datasets used for image deblurring. \citet{hu2014deblurring} introduce a low-illumination dataset for the purpose of recovering low-light blurred images. They capture 11 low-light images and convolve these with 14 different blur kernels to obtain the total of 154 blurred images.  For the application of text recognition, \citet{pan2014deblurring} gather 15 clear document images and apply uniform kernels introduced by \citet{levin2009understanding} to generate blurrred text images.  In addition, \citet{hradivs2015convolutional} provide a large dataset of blurred contents, including text, equations, and tables where the blurred images are generated by applying motion and defocus blurs on the collected text documents. Small geometric transformations with bicubic interpolation are also applied on patches extracted from the dataset to obtain more realistic blurred images. The dataset consists of a total of 3M image patches that can be used for training and 35K patches for testing. There are also other application specific datasets, e.g., face images \cite{huang2008labeled, liu2015deep, kumar2009attribute, sim2002cmu}.  Table~\ref{tab:dataset} summarizes some specifications of the datasets discussed in this section.

\begin{table}[!h]
\begin{center} 
\caption{Image Deblurring Datasets}
\label{tab:dataset}
{\footnotesize
\begin{tabular}{lccccc}
\hline
 Dataset Name  & Domain-specific & Type of dataset & Blur model &Total cases \\
\hline

\citet{levin2009understanding} & General & Synthetic & Uniform & 32 \\
\citet{kohler2012recording} & General & Synthetic & Non-uniform &  48 \\
\citet{sun2013edge} & General & Synthetic & Uniform &640 \\
\citet{hu2014deblurring} & low-illumination & Synthetic & Uniform  & 154\\
\citet{pan2014deblurring}& Text  & Synthetic & Uniform & 120\\
\citet{hradivs2015convolutional} & Text & Synthetic & Non-uniform & 3,035,000\\
\citet{lai2016comparative} & General & Real/Synthetic & Uniform/Non-uniform motion& 300 \\
DeepVideoDeblurring (DVD) \cite{su2017deep} & General & Synthetic & Non-uniform motion & 6,708 \\
GoPro \cite{nah2017deep}  & General  &  Synthetic  & Dynamic scene (multiple sources) & 3,214\\
HIDE \cite{shen2019human} & General & Synthetic & Dynamic scene (multiple sources)& 8,422\\
\citet{rim2020real} & General & Synthetic  & Uniform/Non-uniform motion & 4,738   \\
\hline
\end{tabular} }
\end{center}
\end{table}

\section{Performances of Deep Neural Image Deblurring Methods}
\label{sec:evaluation}
The evaluation of recovered images can be performed by either a qualitative or quantitative manner. Human judgement is more involved in qualitative assessment, for example, by evaluating image clarity, identifying edges, and determining the presence of ringing artifacts.  Yet, quantitative approaches provide a more reliable way to assess image quality by using some metrics that everyone can agree on.  This section discusses widely used metrics in the image deblurring domain for the quantitative assessment of image quality. In addition, we comprehensively list all performance metrics used in the reviewed studies and compare their performance outcomes. 
\subsection{Quantitative Performance Metrics}
In this section, we list popular performance metrics often used in the literature to assess the quality of recovered images. In what follows, suppose $I$ and $\hat{I}$ denote the true image and recovered image, respectively. 
\begin{itemize} 
    \item \textbf{Mean Squared Error (MSE)}\\
    MSE measures the pixel-wise difference between the ground truth and recovered images:
    \begin{equation}
        \text{MSE}(I, \hat{I}) = \frac{1}{N} \sum_{i=1}^N (I_i -  \hat{I}_i)^2 
        \label{mse_met}
    \end{equation}
    where $N$ is the total number of pixels in the image considered. This metric is mostly used for training deep networks rather than evaluating them. When it is used for training, it is similar to the content loss, but it is widely used as a performance metric in the conventional optimization based approaches. The smaller the MSE is, the better the recovery outcome is, that is, the recovered image is more similar to the ground truth image. 
    \item \textbf{Peak Signal-to-Noise Ratio (PSNR)}\\
    PSNR is one of the most widely used metrics in the image deblurring application, and it is formulated as
    \begin{equation}
        \text{PSNR}(I, \hat{I}) = 10  \log \frac{R^2}{\text{MSE}(I, \hat{I})}
        \label{psnr_met}
    \end{equation}
    where $R$ is the maximum possible pixel value for the image. In most cases, images are in an 8-bit format, so $R$ takes a value of 255. The image quality is better when the PSNR has a higher value.  This is a direct consequence of having the MSE in the denominator in Eq.~\eqref{psnr_met} which makes the two metrics inversely proportional \cite{hore2010image}.

    \item \textbf{Structural Similarity Measure (SSIM)} \cite{wang2004image} \\ 
    SSIM is also a very popular metric assessing image quality.  This metric measures the similarity between two images by comparing the patterns of pixel intensities \cite{wang2004image}. Its values range between zero and one, and a higher value indicates a better reconstruction quality. The SSIM is computed as
    \begin{equation}
        \text{SSIM}(I, \hat{I}) = \frac{2\mu_I \mu_{\hat{I}} + C_1}{\mu_I^2 + \mu_{\hat{I}}^2 + C_1}\cdot \frac{2\sigma_{I\hat{I}} + C_2}{\sigma_I^2 + \sigma_{\hat{I}}^2 + C_2}
    \end{equation}
       where $\mu_I / \sigma_I^2$ and $\mu_{\hat{I}} / \sigma_{\hat{I}}^2$ are the mean/variance of pixel values from the true and recovered images, respectively. $C_1$ and $C_2$ are positive constants which are used to stabilize the division.
\end{itemize}

The listed metrics are used widely for general computer vision tasks, especially for image deblurring. There are other proposed metrics that can be used to assess the quality of latent image and estimated blur kernel.  This includes error ratio (ER) \cite{levin2011efficient}, success percent \cite{chakrabarti2016neural}, kernel similarity \cite{hu2015learning}, information fidelity criterion (IFC) \cite{sheikh2005information, ramakrishnan2017deep}, visual information fidelity (VIF) \cite{sheikh2006image, ramakrishnan2017deep}, universal image quality index (UIQI) \cite{wang2002universal, ramakrishnan2017deep}, feature similarity index (FSIM) \cite{zhang2011fsim, ramakrishnan2017deep}, and character error rate (CER) which is typically used for images with text content.

\begin{table}[htbp]
\begin{center} 
\caption{Reported performance of blind deblurring approaches from benchmark datasets, listed in the chronological order.  For the computational time, `h,' `m,' `s,' and `ms' denote hours, minutes, seconds, and miliseconds, respectively.}
\label{tab:perfor}
\resizebox{\textwidth}{!}{
\begin{tabular}{lccccccc}
\toprule
  & & \multicolumn{3}{c}{Performance Measures} \\ \cline{3-5}
  Datasets & Papers & \multicolumn{1}{c}{PSNR (dB)} & \multicolumn{1}{c}{SSIM} & ER Mean/CER & Computational Time \\
  \hline
\midrule
\multirow{3}{*}{\citet{levin2009understanding}} &
\citet{li2019deep} & 27.15 & 0.89 \\&
\citet{ren2020neural} & 33.32 & 0.9438 & 1.2509/-\\&
\citet{zhao2021gradient} & 21.39 & 0.5871 & &  \\
\hline
\multirow{10}{*}{\citet{kohler2012recording}} &
\citet{hyun2013dynamic}   & 24.68 & 0.7937 & &  \\&
\citet{sun2015learning}  & 25.22 & 0.7735 &   & \\&
\citet{ramakrishnan2017deep}  & 27.08 & 0.7510 &  &\\&
\citet{nah2017deep} & 26.48 & 0.8079 &  & \\& 
\citet{kupyn2018deblurgan} & 25.86 & 0.802 & & \\&
\citet{tao2018scale} & 26.75 & 0.837 & &  \\&
\citet{aljadaany2019douglas} & 27.20 & 0.865 &   & \\&
\citet{kupyn2019deblurgan} & 26.72 & 0.836 &  &  \\&
\citet{cai2020dark} & 26.79 & 0.839\\&
\citet{xu2021attentive} & 27.65 & 0.8596 & &   \\& 
\citet{zhao2021gradient} & 26.31 & 0.7858 & &   \\
\hline
\multirow{6}{*}{\citet{sun2013edge}} &
 \citet{schuler2015learning} & $\approx 26.5$ &  &  \\&
 \citet{chakrabarti2016neural} & & & 3.01/-  & 65 s\\&
 \citet{xu2017motion} & $\approx 28$ & \\&
 \citet{nimisha2017blur} & 30.54 & 0.9553 &&3.4 s\\&
 \citet{li2019deep} & 29.91 & 0.93\\&
 \citet{dong2021deep} & 29.69 & 0.9013\\
\hline
\multirow{1}{*}{\citet{pan2014deblurring}} &
\citet{li2018learning} & 28.10 \\
\hline
\citet{hradivs2015convolutional} & \citet{hradivs2015convolutional} & $\approx 24$ & &  -/4$\%$ & \\
\hline
\multirow{2}{*}{\citet{lai2016comparative}} &
\citet{ramakrishnan2017deep} & 27.23 & 0.7651\\&
\citet{ren2020neural} & 21.13 & 0.7319 \\
\hline
\multirow{6}{*}{DeepVideoDeblurring (DVD) \cite{su2017deep}} & 
\citet{zhang2019deep} & 31.43 & - & &   \\&
\citet{kupyn2019deblurgan} & 28.54 & 0.929 & & 0.06 s\\
&  \citet{xu2021attentive} & 31.19 & -\\&
\citet{luo2021bi} & 31.37 & 0.9748 \\&
\citet{li2021single} & 34.64 & 0.960 \\&
\citet{hu2021pyramid} & 31.01 & - & \\
\hline
\multirow{24}{*}{GoPro \cite{nah2017deep}} & 
\citet{hyun2013dynamic} & 23.64 & 0.8239 & & 1 h\\&
\citet{sun2015learning}  & 24.64 & 0.8429 & &  20 m\\ &
\citet{ramakrishnan2017deep} & 28.94 & 0.9220 & &  \\ &
\citet{nah2017deep} & 29.08 & 0.9135 & & 3.09 s \\ &
\citet{kupyn2018deblurgan} & 28.7  & 0.958  & &  0.85 s\\ &
\citet{zhang2018dynamic} & 29.187 & 0.9306 & & 1.4 s\\ &
\citet{tao2018scale} & 30.26 & 0.9342  & & 1.87 s \\ &
\citet{zhang2019deep} & 31.20 & 0.9453 & & 0.042 s \\&
\citet{aljadaany2019douglas}  & 30.35 & 0.961 & & 1.2 s\\ &
\citet{kupyn2019deblurgan} & 29.55 & 0.934 & & 0.35 s\\&
\citet{gao2019dynamic} & 30.92 & 0.9421 & & 1.6 s\\ &
\citet{shen2019human} & 30.26 & 0.940 &&\\&
\citet{cai2020dark} & 31.10 & 0.945 && 0.65 s\\&
\citet{purohit2020region} & 31.76 & 0.9530 && 38 ms \\&
\citet{zhang2020deblurring} & 31.10 & 0.9424 && \\&
\citet{xu2021attentive} & 31.23 & 0.9455 & &  0.28 s\\&  
\citet{zhao2021gradient} & 30.67 & 0.9372 && 0.598 s \\&
\citet{chen2021attention} & 31.34 & 0.9467 && 30 ms\\& 
\citet{cho2021rethinking} & 32.68 & 0.959 & & 0.040 s\\& 
\citet{luo2021bi} & 30.18 & 0.9569 && 0.09 s \\& 
\citet{li2021single} & 30.21 & 0.905 & & 1.05 s\\& 
\citet{ren2021deblurring} & 30.46 & 0.9365 & & 1.4 s\\&
\citet{hu2021pyramid} & 30.62 & 0.9405 && 17 ms \\&
\citet{tsai2021banet} & 32.44 & 0.957 && 28 ms \\
\hline
\multirow{2}{*}{HIDE \cite{shen2019human}} & 
\citet{shen2019human} & 29.60 & 0.941 &&\\&
\citet{tsai2021banet} & 30.27 & 0.931 && 26 ms\\
\hline
\citet{rim2020real} & 
\citet{cho2021rethinking} & 31.73 &  \\
\bottomrule
\end{tabular}
}
\end{center}
\end{table}


\subsection{Performance Comparison}
\par 
Table \ref{tab:perfor} summarizes and compares the performance of the deep neural image deblurring approaches reviewed in this paper with respect to the metrics discussed in the previous section.  Since these studies conduct experiments by using the benchmark datasets listed in Section~\ref{sec:deblur_dataset}, some of them can be directly compared.  In addition, we add computational time for the GoPro dataset \cite{nah2017deep} where the time is measured for recovering a single image of size 720$\times$1280. 
The table clearly shows how frequently each dataset has been used and how the performance of the proposed studies evolves over time while reducing the computational time.

%

\section{Challenges and Future Directions}
\label{sec:challenge}
\par
Image deblurring is still a challenging research topic in computer vision, and deep learning-based approaches start gaining popularity just recently. In this section, we discuss the current challenges of deep neural image deblurring methods and provide some possible directions for future works. 

\begin{itemize}
\item \textbf{Architecture scalability and generalizability} \\
     The current deep neural image deblurring architectures are lack of scalability and generalizability.  For deep learning structures which require extensive training, improving model scalability is crucial for their usage in various applications.  In the presence of massive computational requirements, training a model up to a certain accuracy itself can be very time consuming, and hence it is hard to expect such a model can be adapted for applications to other problems.  
     There are mainly three aspects to consider: the size and complexity of a model, the volume of training datasets, and the specifications of hardware \cite{mayer2020scalable}, e.g., using GPU for the training step. The first two aspects and their future directions are discussed in more detail in the following bullet points, including feature extraction, architecture complexity, and image deblurring dataset. Concerning generalizability, the current deblurring architectures are not quite adaptive to various applications. That is,   
     some general architectures would perform poorly on some specific applications, such as face image and text images, since these domains have their own distinct characteristics. In this context, future architectures should consider some semantic information as well as inherent features to build more generalizable structures. 
    \item \textbf{Feature extraction}\\
    Increasing the depth of a network structure does not necessarily improve the quality of deblurring outcomes \cite{zhang2019deep}. For this reason, an effective extraction of inherent features is very critical in a deblurring process, and this suggests a need for innovative modules that can effectively extract all the beneficial information. Although the up-sampling and down-sampling operations in multi-scale architecture is developed to extract more information from an image in varying scales, it weakens the importance of resolution in each scale, without fully utilizing high-frequency contents that are important for image deblurring \cite{purohit2020region}.
    An integration of intelligent feature extraction modules can help retrieve constructive information for a more in-depth deblurring procedure without making the network itself far deeper. Recently, various attention modules, pyramid scales, weight sharing schemes, and feature extraction blocks are developed for the encoder section of a network structure; however, some proper combination of these modules has a potential to acquire more useful information restoring higher quality latent images and thus is worth studying. 
    \item \textbf{Architecture complexity}\\
    The architecture complexity is the major component affecting the run time and required memory of deep neural image deblurring architectures. For instance, an addition of more convolutional layers or the upsampling strategy in a multi-scale mechanism dramatically increases required computations \cite{zhang2019deep}. The former structure naturally requires more convolution operations with more layers, and the latter requires to include scale-independent weights that should be optimized during the training process. Structure stacking, shared weight schemes, and deep enough single-scale architecture can significantly diminish computational cost so will be a good candidate for future development of deep neural image deblurring networks. 
    \item \textbf{Training loss functions}\\
    The selection of a loss function dictates the effectiveness of the training process and thereby the quality of image recovery. As shown in Table \ref{tab:loss}, the studies in the literature propose and perform ablation study for different types of loss functions to achieve the best performance. In general, fusing proper loss functions can improve the model performance, but which loss functions to combine for particular applications needs more studies.  In addition, developing a loss function that works well for a broad set of applications is very challenging, which requires more verification and evaluation.
    \item \textbf{Image deblurring Datasets} \\
    The image deblurring datasets currently available in the literature include synthetic pairs of blurry/sharp images.  As a consequence, trained networks often perform poorly on some real blurry images \cite{zhang2022deep}. To address this issue, some efforts need to follow to study real-world blurring effects and blurring sources and capture massive realistic images based on the understanding.  On the other hand,
    well-trained deep learning structures can be used to generate more realistic blurred images, for example, as has been done by \citet{zhang2020deblurring} where the network is trained to make synthetic blurred images that are indistinguishable from real blurry images. 
\end{itemize}

\section{Conclusion}
\label{sec:conclusion}
This paper reviews the deep neural image deblurring studies and describes their advances since the initial introduction of the concept. The most widely used deep elements and popular deblurring mechanisms are initially described. A comprehensive review of deep neural image deblurring methods follows afterward, which includes non-blind and blind deblurring approaches, deep learning-based image priors, and specific applications structures. Furthermore, the key components of individual deblurring architectures along with the corresponding loss functions, their applications, and blur types are thoroughly explained in this paper. The most popular deblurring datasets are outlined, and a quantitative performance comparison of the reviewed papers is provided to highlight the impact of each structure on the quality of the recovered images. This paper also discusses the current challenges in deep neural image deblurring and provides some guidance for future studies.  

\bibliography{review_refs.bib} 
\bibliographystyle{plainnat}

\end{document}